\documentclass[10pt,journal,compsoc]{IEEEtran}
\usepackage{graphicx}
\usepackage{amsfonts}
\usepackage{booktabs}
\usepackage{multirow}
\usepackage{subfigure}
\usepackage{framed}
\usepackage{color}

\ifCLASSOPTIONcompsoc
  \usepackage[nocompress]{cite}
\else
  \usepackage{cite}
\fi
\ifCLASSINFOpdf
\else
\fi
\begin{document}
%
\title{Spatio-Temporal Joint Graph Convolutional Networks for Traffic Forecasting}
%
%
%
%

\author{Chuanpan Zheng,
        Xiaoliang Fan,~\IEEEmembership{Senior Member,~IEEE,}
        Shirui Pan,~\IEEEmembership{Senior Member,~IEEE,} 
        Haibing Jin, \\
        Zhaopeng Peng,
        Zonghan Wu, 
        Cheng Wang,~\IEEEmembership{Senior Member,~IEEE,} 
        and~Philip S. Yu,~\IEEEmembership{Fellow,~IEEE,} 
\IEEEcompsocitemizethanks{\IEEEcompsocthanksitem C. Zheng, X. Fan, H. Jin, Z. Peng and C. Wang are with Fujian Key Laboratory of Sensing and Computing for Smart Cities, School of Informatics, Computer Science and Technology Department, and Key Laboratory of Multimedia Trusted Perception and Efficient Computing, Ministry of Education of China, Xiamen University, Xiamen, 361005, China.\protect\\
E-mail: zhengchuanpan@stu.xmu.edu.cn; fanxiaoliang@xmu.edu.cn; jinhaibing@stu.xmu.edu.cn; pengzhaopeng@stu.xmu.edu.cn; cwang@xmu.edu.cn
\IEEEcompsocthanksitem S. Pan is with School of Information and Communication Technology, Griffith University, Australia.\protect\\
E-mail: s.pan@griffith.edu.au
\IEEEcompsocthanksitem Z. Wu is with Centre for Artificial Intelligence, FEIT, University of Technology Sydney, Australia.\protect\\
E-mail: zohn.wu@gmail.com
\IEEEcompsocthanksitem P. S. Yu is with the Department of Computer Science, University of Illinois at Chicago, Chicago, IL 60607 USA.\protect\\
E-mail: psyu@cs.uic.edu.}
\thanks{(Corresponding author: Xiaoliang Fan)}}

%
%

\markboth{IEEE TRANSACTIONS ON KNOWLEDGE AND DATA ENGINEERING}%
{Shell \MakeLowercase{\textit{et al.}}: Bare Demo of IEEEtran.cls for Computer Society Journals}
%



\IEEEtitleabstractindextext{%
\begin{abstract}
	Recent studies have shifted their focus towards formulating traffic forecasting as a spatio-temporal graph modeling problem. Typically, they constructed a static spatial graph at each time step and then connected each node with itself between adjacent time steps to create a spatio-temporal graph. However, this approach failed to explicitly reflect the correlations between different nodes at different time steps, thus limiting the learning capability of graph neural networks. Additionally, those models overlooked the dynamic spatio-temporal correlations among nodes by using the same adjacency matrix across different time steps. To address these limitations, we propose a novel approach called Spatio-Temporal Joint Graph Convolutional Networks (STJGCN) for accurate traffic forecasting on road networks over multiple future time steps. Specifically, our method encompasses the construction of both pre-defined and adaptive spatio-temporal joint graphs (STJGs) between any two time steps, which represent comprehensive and dynamic spatio-temporal correlations. We further introduce dilated causal spatio-temporal joint graph convolution layers on the STJG to capture spatio-temporal dependencies from distinct perspectives with multiple ranges. To aggregate information from different ranges, we propose a multi-range attention mechanism. Finally, we evaluate our approach on five public traffic datasets and experimental results demonstrate that STJGCN is not only computationally efficient but also outperforms 11 state-of-the-art baseline methods.
\end{abstract}

}

\maketitle

\IEEEdisplaynontitleabstractindextext

%
\IEEEpeerreviewmaketitle

\IEEEraisesectionheading{\section{Introduction} \label{Introduction}}

%
%
%
%

\IEEEPARstart{S}{patio-temporal} data forecasting has received increasing attention from the deep learning community in recent years~\cite{Wang-et-al:TKDE2020,Tedjopurnomo-et-al:TKDE2020,Zheng-et-al:TITS2021}. It plays a vital role in a wide range of applications, such as traffic speed prediction~\cite{Li-et-al:ICLR2018} and air quality inference~\cite{Cheng-et-al:AAAI2018}. In this paper, we study the problem of forecasting the future traffic conditions given historical observations on a road network.

\begin{figure*}
	\centering
	\subfigure[Spatio-Temporal Graph]{
		\label{fig1(a)} 
		\includegraphics[width = 0.23 \textwidth, height = 0.18 \textwidth]{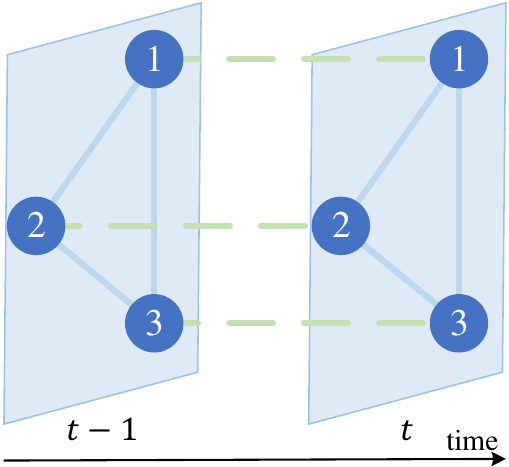}}
		\hspace{10pt}		
	\subfigure[Pre-defined STJG]{
		\label{fig1(b)} 
		\includegraphics[width = 0.23 \textwidth, height = 0.18 \textwidth]{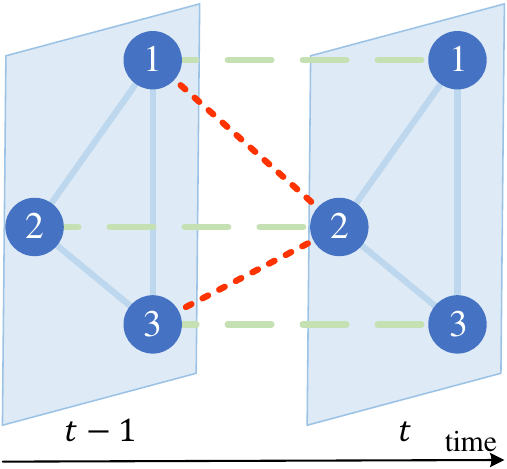}}
		\hspace{10pt}
	\subfigure[Adaptive STJG]{
		\label{fig1(c)} 
		\includegraphics[width = 0.23 \textwidth, height = 0.18 \textwidth]{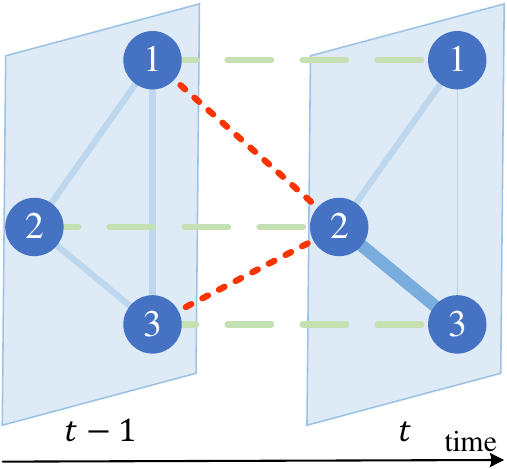}}
	\subfigure{
		\includegraphics[width = 0.5 \textwidth]{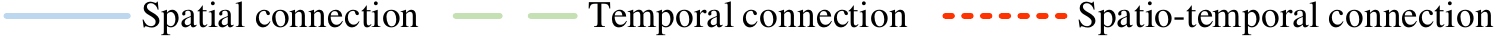}}
	\caption{The comprehensive and dynamic connections among nodes in graph-structured spatio-temporal data. There are three common scenarios: (a) Spatio-Temporal Graph: The node 2 at time step $ t $ can be influenced by nodes 1 and 3 at time step $ t $ through spatial connections, and node 2 at time step $ t-1 $ through the temporal connection. (b) Pre-defined Spatio-Temporal Joint Graph: The node 2 at time step $ t $ may also be affected by nodes 1 and 3 at time step $ t-1 $ through spatio-temporal connections. (c) Adaptive Spatio-Temporal Joint Graph: Compared with time step $ t-1 $, the connections among nodes 1, 2 and 3 exhibit strong dynamic characteristics at the time step $ t $. For instance, the connection between nodes 1 and 3 gets weakened, while the connection between nodes 2 and 3 becomes stronger. Both (b) and (c) scenarios have not been comprehensively explored in existing studies.}
	\label{fig1}
\end{figure*}

Recent studies formulate traffic forecasting as a spatio-temporal graph modeling problem~\cite{Li-et-al:ICLR2018,Yu-et-al:IJCAI2018,Wu-et-al:IJCAI2019,Guo-et-al:AAAI2019,Song-et-al:AAAI2020,Bai-et-al:NIPS2020, Chen-et-al:ICML2021}. The basic assumption is that the state of each node is conditioned on its neighboring node information. Based on this, they construct a spatial graph with a pre-defined~\cite{Li-et-al:ICLR2018} or data-adaptive~\cite{Wu-et-al:IJCAI2019} adjacency matrix. In such a graph, each node corresponds to a location of interest (e.g., traffic sensor). The graph neural network~\cite{Wu-et-al:TNNLS2021} is applied on that graph to model the correlations among spatial neighboring nodes at each time step. To leverage the information from temporal neighboring nodes, they further connect each node with itself between adjacent time steps, which results in a spatio-temporal graph, as shown in Figure~\ref{fig1(a)}. The 1D convolutional neural network~\cite{Yu-et-al:IJCAI2018} or recurrent neural network~\cite{Li-et-al:ICLR2018} is commonly used to model the correlations at each node between different time steps. By combining the spatial and temporal features, they are able to update the state of each node. 

However, those spatio-temporal graphs do not explicitly reflect the correlations between different nodes at different time steps (e.g., the red dash lines in Figure~\ref{fig1(b)}). In such a graph, the information of spatial and temporal neighborhoods is captured through the spatial and temporal connections respectively, while the information of neighboring nodes across both spatial and temporal dimensions are not considered, which may restrict the learning ability of graph neural networks. For example, a traffic jam occurred at an intersection may affect not only current nearby roads (spatial neighborhoods) and its local future traffic condition (temporal neighborhoods), but also the downstream roads in next few hours (spatio-temporal neighborhoods). Thus, we argue that it is necessary to model the comprehensive correlations in the spatio-temporal data.

Another limitation of previous works is that they ignore the dynamic correlations among nodes at different time steps, as shown in Figure~\ref{fig1(c)}. The road network distances among sensors (nodes) are commonly used to define the spatial graph~\cite{Li-et-al:ICLR2018,Yu-et-al:IJCAI2018}. This pre-defined graph is usually static. Some researchers~\cite{Wu-et-al:IJCAI2019,Bai-et-al:NIPS2020} propose to learn a data-adaptive adjacency matrix, which is also unchanged over time steps. However, the traffic data exhibits strong dynamic correlations in the spatial and temporal dimensions, those static graphs are unable to reflect the dynamic characteristics of correlations among nodes. For example, the residence region is highly correlated to the office area during workday morning rush hours, while the correlation would be relatively weakened in the evening because some people might prefer to dining out before going home. Thus, it is crucial to model the dynamic spatio-temporal correlations for traffic forecasting.

This paper addresses these limitations from the following perspectives. First, besides the spatial and temporal connections, we further add the spatio-temporal connections between two time steps according to the spatio-temporal distances to define the \textit{spatio-temporal joint graph} (STJG). In this way, the pre-defined STJG preserves comprehensive spatio-temporal correlations between any two time steps. Second, in order to adapt to the dynamic correlations among nodes, we suggest to explore an adaptive STJG, which is time-variant by encoding the time features. The adjacency matrix in this adaptive STJG is dynamic, changing over time steps. By constructing both the pre-defined and adaptive STJGs, we are able to preserve comprehensive and dynamic spatio-temporal correlations.


On these basis, we then develop the~\textit{spatio-temporal joint graph convolution} (STJGC) operations on both pre-defined and adaptive STJGs to simultaneously capture the spatio-temporal dependencies in a unified operation. We further design the dilated causal STJGC layers to extract multiple spatio-temporal ranges of information. Next, a multi-range attention mechanism is proposed to aggregate the information of different ranges. Finally, we apply independent fully-connected layers to produce the multi-step ahead prediction results. The whole framework is named as \textit{spatio-temporal joint graph convolutional networks} (STJGCN), which can be learned end-to-end. To evaluate the efficiency and effectiveness of STJGCN, we conduct extensive experiments on five public traffic datasets. The experimental results demonstrate that our STJGCN is computationally efficient and achieves the best performance against 11 state-of-the-art baseline methods. Our main contributions are summarized as follows.

\begin{itemize}
	\item We construct both pre-defined and adaptive spatio-temporal joint graphs (STJGs), which reflect comprehensive and dynamic spatio-temporal correlations. 
	\item We design dilated causal spatio-temporal joint graph convolution layers on both types of STJG to model multiple ranges of spatio-temporal correlations.
	\item We propose a multi-range attention mechanism to aggregate the information of different ranges. 
	\item We evaluate our model on five public traffic datasets, and experimental results demonstrate that STJGCN has high computation efficiency and outperforms 11 state-of-the-art baseline methods.
\end{itemize}

The rest of this paper is organized as follows. Section~\ref{Related Work} reviews the related work. Section~\ref{Preliminary} presents the preliminary of this work. Section~\ref{Methodology} details the method of STJGCN. Section~\ref{Experiments} compares STJGCN with state-of-the-art methods on five datasets. Finally, section~\ref{Conclusion} concludes this paper and draws future work.

\section{Related Work} \label{Related Work}

\subsection{Graph Convolutional Networks} 

Graph convolutional networks (GCNs) are successfully applied on various tasks (e.g., node classification~\cite{Kipf-and-Welling:ICLR2017}, link prediction~\cite{Zhang-et-al:NIPS2018}) due to their superior abilities of handling graph-structured data~\cite{Wu-et-al:TNNLS2021}. There are mainly two types of GCN~\cite{Bronstein-et-al:SPM2017}: spatial GCN and spectral GCN. The spatial GCN performs convolution filters on neighborhoods of each node. Researchers in~\cite{Niepert-et-al:ICML2016} propose a heuristic linear method for neighborhood selecting. GraphSAGE~\cite{Hamilton-et-al:NIPS2017} samples a fixed number of neighbors for each node and aggregates their features. GAT~\cite{Velickovic-et-al:ICLR2018} learns the weights among nodes via attention mechanisms. Researchers in~\cite{jia2020residual} improve graph neural network architecture by exploiting correlation structure in the regression residuals. The spectral GCN defines the convolution in the spectral domain~\cite{Li-et-al:AAAI2020}, which is firstly introduced in~\cite{Bruna-et-al:ICLR2014}. ChebNet~\cite{Defferrard-et-al:NIPS2016} reduces the computational complexity with fast localized convolution filters. In~\cite{Kipf-and-Welling:ICLR2017}, researchers further simplify the ChebNet to a simpler form and achieve state-of-the-art performances on various tasks. Recently, a range of studies apply the GCN on time-series data and construct spatio-temporal graphs for traffic forecasting~\cite{Li-et-al:ICLR2018,Zhang-et-al:TITS2021}, human action recognition~\cite{Yan-et-al:AAAI2018,Shi-et-al:CVPR2019}, etc. 

\subsection{Spatio-Temporal Forecasting} 

Spatio-temporal forecasting is an important research topic, which has been extensively studied for decades~\cite{Yin-et-al:arXiv2020,Sun-et-al:TKDE2020,Gu-et-al:TKDE2020,Jiang-et-al:TKDE2021,Guo-et-al:TKDE2021}. Recurrent neural networks (RNNs), especially the long short-term memory (LSTM) and gated recurrent unit (GRU) are successfully applied for modeling temporal correlations~\cite{Ma-et-al:TRC2015}. To capture the spatial dependencies, convolutional neural networks (CNNs) are introduced, which are restricted to process regular grid structures~\cite{Zhang-et-al:AAAI2017,Yao-et-al:AAAI2018,Yao-et-al:AAAI2019,Zheng-et-al:TITS2020,Zhang-et-al:TKDE2020}. Recently, researchers apply graph neural networks to model the non-Euclidean spatial correlations~\cite{Ye-et-al:arXiv2020}. DCRNN~\cite{Li-et-al:ICLR2018} employs diffusion convolution to capture the spatial dependency and applies GRU to model the temporal dependency. STGCN~\cite{Yu-et-al:IJCAI2018} uses graph convolution and 1D convolution to model the spatial and temporal dependencies, respectively. Researchers in~\cite{lau2021spatio} study the effect of the order of spatial layers and temporal layers on STGCN model performance. Several works~\cite{Guo-et-al:AAAI2019,Zheng-et-al:AAAI2020,Wang-et-al:WWW2020} introduce the attention mechanisms~\cite{Vaswani-et-al:NIPS2017} into the spatio-temporal graph modeling to improve the prediction accuracy. 
AGSTN~\cite{lu2020agstn} proposes an attention adjustment mechanism to realize fluctuation modulation for learning time-evolving spatio-temporal correlation.
Some studies consider more kinds of connections (e.g., semantic connection~\cite{Wang-et-al:KDD2019}, edge interaction patterns~\cite{Chen-et-al:AAAI2020}) to construct the spatial graph. The adjacency matrices in these models are usually pre-defined according to some prior knowledge (e.g., distances among nodes). Some researchers~\cite{Wu-et-al:IJCAI2019,Bai-et-al:NIPS2020} argue that the pre-defined adjacency matrix does not necessarily reflect the underlying dependencies among nodes, and propose to learn an adaptive adjacency matrix for graph modeling. However, both the pre-defined and adaptive adjacency matrices assume static correlations among nodes, which cannot adapt to the evolving systems (e.g., traffic networks). Moreover, these graph-based methods do not explicitly model the correlations between different nodes at different time steps, which may restrict the learning ability of graph neural networks.  

\section{Preliminary} \label{Preliminary}

\textit{Problem definition.} Suppose there are $ N $ sensors (nodes) on a road network, and each sensor records $ C $ traffic measurements (e.g., volume, speed) at each time step. Thus, the traffic conditions at time step $ t $ can be represented as $ X_{t} \in \mathbb{R}^{N \times C} $.  The traffic forecasting problem aims to learn a function $ f $ that maps the traffic conditions of historical $ P $ time steps to next $ Q $ time steps:
%
\begin{equation}
[X_{t-P+1},X_{t-P+2},\cdots,X_{t}]\stackrel{f}{\longrightarrow}[X_{t+1},X_{t+2},\cdots,X_{t+Q}].
\label{eq1}
\end{equation}
\section{Methodology} \label{Methodology}

\begin{figure*}
	\centering
	\includegraphics[width = 0.8 \textwidth, height = 0.3 \textwidth]{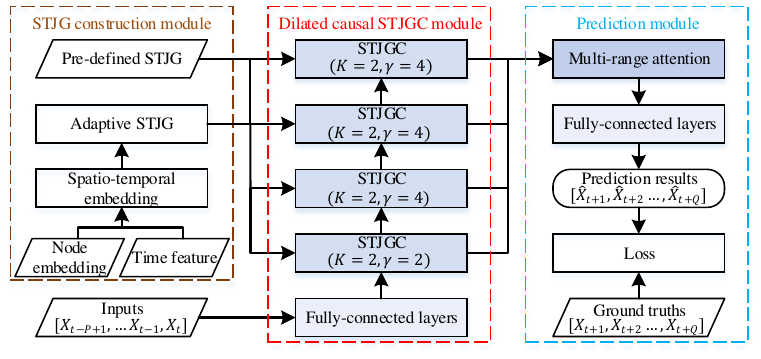}
	\caption{The framework of \textit{Spatio-Temporal Joint Graph Convolutional Networks} (STJGCN). It consists of three modules: (i) the \textit{STJG construction module} (detailed in section~\ref{STJG Construction Module}) constructs both pre-defined and adaptive \textit{spatio-temporal joint graphs} (STJGs); (ii) the \textit{dilated causal STJGC module} (detailed in section~\ref{Dilated Causal STJGC Module}) stacks dilated causal \textit{spatio-temporal joint graph convolution} (STJGC) layers to capture multiple ranges of spatio-temporal dependencies, where each STJGC layer performs convolution operation based on both types of STJG; (iii) the \textit{prediction module} (detailed in section~\ref{Prediction Module}) aggregates the information of different ranges via a multi-range attention mechanism and produces the prediction results using fully-connected layers.}
	\label{fig2}
\end{figure*}

\subsection{Framework Overview} 

Figure~\ref{fig2} depicts the framework of our proposed Spatio-Temporal Joint Graph Convolutional Networks (STJGCN), which includes three modules. First, previous graph-based methods generally ignore the spatio-temporal connections and the dynamic correlations among nodes, we thus propose the \textit{spatio-temporal joint graph (STJG) construction module} to construct both pre-defined and adaptive STJGs, which preserve comprehensive and dynamic spatio-temporal correlations. Second, as the standard graph convolution operation models spatial correlations only, we propose the \textit{spatio-temporal joint graph convolution} (STJGC) operation on both types of STJG to model the comprehensive and dynamic spatio-temporal correlations in a unified operation. Based on the STJGC, we further propose the \textit{dilated casual STJGC module} to capture spatio-temporal dependencies within multiple neighborhood and time ranges. Finally, in the \textit{prediction module}, we propose a multi-range attention mechanism to aggregate the information of different ranges, and apply fully-connected layers to produce the prediction results. We detail each module in the following subsections.

\subsection{STJG Construction Module} \label{STJG Construction Module}

In this module, we first pre-define the \textit{spatio-temporal joint graph} (STJG) according to the spatio-temporal distances among nodes. While, the pre-defined graph may not reflect the underlying correlations among nodes~\cite{Wu-et-al:IJCAI2019,Bai-et-al:NIPS2020}, we further propose to learn adaptive STJG. By constructing both types of STJG, we are able to represent comprehensive and dynamic spatio-temporal correlations among nodes.

\subsubsection{Pre-defined Spatio-Temporal Joint Graph}

Previous studies~\cite{Li-et-al:ICLR2018,Yu-et-al:IJCAI2018} for traffic forecasting on graphs usually define the spatial adjacency matrix based on pair-wise road network distances:
\begin{equation}
A_{i,j}=\exp(-\frac{{dist(v_{i},v_{j})}^2}{\sigma^2})
\label{eq2},
\end{equation}
where $ dist(v_{i},v_{j}) $ represents the road network distance from node $ v_{i} $ to node $ v_{j} $, $ \sigma $ is the standard deviation of distances, and $ A_{i,j} $ denotes the edge weight between node $ v_i $ and node $ v_j $. They construct the spatial graph at each time step, and then connect each node with itself between adjacent time steps to define the spatio-temporal graph. In such a graph, the connections between different nodes at different time steps are not incorporated, which may restrict its representation ability. 

We propose to construct a \textit{spatio-temporal joint graph} (STJG), which preserves comprehensive spatio-temporal correlations. The intuitive idea is to further connect different nodes between two time steps, as shown in Figure~\ref{fig1(b)}. Thus, we modify Equation~\ref{eq2} to be the STJG adjacency matrix, as:
\begin{equation}
A_{i,t-k;j,t}=\exp(-\frac{{((k + 1) \cdot dist(v_{i},v_{j}))}^2}{\sigma^2})
\label{eq3},
\end{equation}
where $ k $ is the time difference between two time steps. $ A_{i,t-k;j,t} $ defines the edge weight between node $ v_i $ at time step $ t-k $ and node $ v_j $ at time step $ t $, which decreases with the increase of spatio-temporal distance. When $ k=0 $, Equation~\ref{eq3} degenerates to Equation~\ref{eq2}, which represents the spatial connections. If $ i=j $, the STJG adjacency matrix defines the temporal connections at each node between two time steps. Otherwise, it represents the spatio-temporal connections between different nodes at different time steps. Thus, we are able to define a comprehensive spatio-temporal graph according to~Equation~\ref{eq3}. Note that the STJG could be constructed between any two time steps, which makes it flexible to reveal multiple time-ranges of spatio-temporal correlations. 

We filter the values smaller than a threshold $ \delta_{pdf} $ in the STJG adjacency matrix to eliminate weak connections and control the sparsity. As this adjacency matrix is conditioned on the time difference $ k $, but irrelevant to a specific time step, we denote it as $ A^{(k)} \in \mathbb{R}^{N \times N} $ in following discussions. 

\subsubsection{Adaptive Spatio-Temporal Joint Graph}

Previous studies~\cite{Wu-et-al:IJCAI2019,Bai-et-al:NIPS2020} demonstrate that the pre-defined adjacency matrix may not reflect the underlying correlations among nodes, and propose adaptive ones. However, they only define the spatial graph, and it is unchanged over time steps. 
We propose to learn adaptive STJG adjacency matrices that could represent comprehensive and dynamic spatio-temporal correlations based on the latent space modeling algorithm~\cite{Deng-et-al:KDD2016}. 

\paragraph{\textit{Latent space modeling}} Given a graph, we assume each node resides in a latent space with various attributes. The attributes of nodes and how these attributes interact with each other jointly determine the underlying relations among nodes. The nodes which are close to each other in the latent space are more likely to form a link. Mathematically, we aim to learn two matrices $ U $ and $ B $. Here, $ U \in \mathbb{R}^{N \times d} $ denotes the $ d $ latent attributes of the $ N $ nodes, and $ B \in \mathbb{R}^{d \times d} $ represents the attributes interaction patterns, which could be an asymmetric matrix for directed graph or symmetric matrix for undirected graph. The product of $ UBU^\top $ could represent the connections among nodes. 

\paragraph{\textit{Spatio-temporal embedding}} We propose a spatio-temporal embedding to form the latent node attributes. We first randomly initialize a spatial embedding for each of the $ N $ nodes, and then transform it to $ d $ dimensions via fully-connected layers. To obtain time-varying node attributes and take periodic patterns in historical input data (i.e., morning rush hour) into account, we further encode the time information as the temporal embedding. At each time step, we consider two time features, i.e., time-of-day and day-of-week, which are encoded by one-hot coding and then be projected to $ d $ dimensions using fully-connected layers. We then add the spatial and temporal embeddings together to generate the spatio-temporal embedding at each time step $ t $, represented as $ U_t \in \mathbb{R}^{N \times d} $, which can be updated during the training stage. The spatio-temporal embedding encodes both the node-specific and time-varying information, and it could mine periodic spatio-temporal patterns of historical data. 

\paragraph{\textbf{\textit{Adaptive STJG adjacency matrix}}} Based on the spatio-temporal embedding, we define the STJG adjacency matrix at time step $ t $ according to the latent space modeling algorithm, as:
\begin{equation}
\tilde{L}_{t}=softmax(\psi(U_tBU_t^\top)) 
\label{eq4},
\end{equation}
with
\begin{equation}
\psi(x) = \left\{ 
\begin{array}{lr}
x,~~~~if~~x \ge \delta_{adt} \\
0,~~~~otherwise 
\end{array}
\label{eq5},
\right.
\end{equation}
where $ U_t \in \mathbb{R}^{N \times d} $ is the spatio-temporal embedding of $ N $ nodes at time step $ t $, $ \psi(x) $ is used to eliminate the weights smaller than a threshold $ \delta_{adt} $, and the softmax function is applied for normalization. $ \tilde{L}_{t} \in \mathbb{R}^{N \times N} $ defines the spatial connections among $ N $ nodes at time step $ t $, which is dynamic, changing over time steps. In order to construct the connections between different time steps, we modify Equation~\ref{eq4} as:
\begin{equation}
\tilde{L}_{t-k;t}=softmax(\psi(U_{t-k}BU_t^\top))
\label{eq6},
\end{equation}
where $ \tilde{L}_{t-k;t} \in \mathbb{R}^{N \times N} $ is the normalized STJG adjacency matrix between time steps $ t-k $ and $ t $. When $ k=0 $, Equation~\ref{eq6} degenerates to Equation~\ref{eq4}, which describes the spatial graph at time step $ t $. Thus, Equation~\ref{eq6} is able to define the spatio-temporal joint graph between time steps $ t-k $ and $ t $ with comprehensive and dynamic spatio-temporal connections.

\subsection{Dilated Causal STJGC Module} \label{Dilated Causal STJGC Module}

The standard graph convolution performs on spatial graphs to model spatial correlations only, we thus propose the \textit{spatio-temporal joint graph convolution} (STJGC) on both types of STJG to model spatio-temporal correlations in a unified operation. We further design dilated causal STJGC layers to capture multiple ranges of spatio-temporal dependencies, as shown in Figure~\ref{fig2}. In the following discussion, we first describe the STJGC operation in section~\ref{Spatio-Temporal Joint Graph Convolution (STJGC)}, and then introduce the dilated causal STJGC layers in section~\ref{Dilated Causal STJGC Layers}.  

\subsubsection{Spatio-Temporal Joint Graph Convolution (STJGC)} \label{Spatio-Temporal Joint Graph Convolution (STJGC)}

Graph convolution is an effective operation for learning node information from spatial neighborhoods according to the graph structure, while the standard graph convolution performs on the spatial graph to model the spatial correlations only. In order to model the comprehensive and dynamic spatio-temporal correlations on the STJG, we propose the spatio-temporal joint graph convolution (STJGC) operations on both types of STJG.

\paragraph{\textit{Graph Convolution}} The graph convolution is defined as~\cite{Kipf-and-Welling:ICLR2017}:
\begin{equation}
Z=\phi(\tilde{A}XW+b).
\label{eq7}
\end{equation}
Here, $ X \in \mathbb{R}^{N \times d_1} $ and $ Z \in \mathbb{R}^{N \times d_2} $ denote the input and output graph signals, $ W \in \mathbb{R}^{d_1 \times d_2} $ and $ b \in \mathbb{R}^{d_2} $ are learnable parameters, $ \phi(\cdot) $ is an activation function (e.g., ReLU~\cite{Nair-and-Hinton:ICML2010}), $ \tilde{A}=D^{-1/2}AD^{-1/2} \in \mathbb{R}^{N \times N} $ is the normalized adjacency matrix, where $ A $ is the adjacency matrix with self-loops, and $ D=\sum_{j}A_{i, j} $ is the degree matrix.

\paragraph{\textbf{\textit{STJGC on pre-defined STJG}}}
Consider the STJG between time steps $ t-k $ and $ t $, the information of each node at time step $ t $ comes from its spatial, temporal, and spatio-temporal neighborhoods:
\begin{equation}
Z_t^{pdf}=\phi(\tilde{A}^{(k)}X_{t-k}W_1^{pdf}+\tilde{A}^{(0)}X_tW_2^{pdf}+b^{pdf}),
\label{eq8}
\end{equation}
where $ \tilde{A}^{(k)} $ is the normalized pre-defined STJG adjacency matrix between time steps $ t-k $ and $ t $ (see Equation~\ref{eq3}). In Equation~\ref{eq8}, $ \tilde{A}^{(k)}X_{t-k}W_1^{pdf} $ means we aggregate neighborhoods (both temporal and spatio-temporal) information from time step $ t-k $, and $ \tilde{A}^{(0)}X_tW_2^{pdf} $ means we aggregate the information from spatial neighborhoods at time step $ t $. Thus, by performing Equation~\ref{eq8}, we are able to model comprehensive spatio-temporal correlations between two time steps.

Furthermore, at time step $ t $, we propose to incorporate $ K $ (denoted as kernel size) time step information (e.g., $ t,t-1,\cdots,t-K+1 $) to update the node features. Specifically, we modify Equation~\ref{eq8} as: 
\begin{equation}
Z_t^{pdf}=\sum_{k=0}^{K-1}\phi(\tilde{A}^{(k)}X_{t-k}W_k^{pdf}+b^{pdf}).
\label{eq9}
\end{equation}

In the case of a directed graph, we consider two directions of information propagation (i.e., forward and backward), corresponding to two normalized adjacency matrices: $ \tilde{A}_{fw}^{(k)}={D_{O}^{(k)}}^{-1/2}A^{(k)}{D_{O}^{(k)}}^{-1/2} $ and $ \tilde{A}_{bw}^{(k)}={D_{I}^{(k)}}^{-1/2}{A^{(k)}}^\top{D_{I}^{(k)}}^{-1/2} $, where $ D_O^{(k)}= \sum_{j}A_{i,j}^{(k)} $ and $ D_I^{(k)}= \sum_{i}A_{i,j}^{(k)} $ represent the out-degree and in-degree matrices, respectively. Thus, we transform Equation~\ref{eq9} to:
%
%
%
\begin{equation}
Z_t^{pdf}=\sum_{k=0}^{K-1}\phi(\tilde{A}_{fw}^{(k)}X_{t-k}W_{k,1}^{pdf}+\tilde{A}_{bw}^{(k)}X_{t-k}W_{k,2}^{pdf}+b^{pdf}),
\label{eq10}
\end{equation}
where $ X_{t-k} \in \mathbb{R}^{N \times d} $ and $ X_{t} \in \mathbb{R}^{N \times d} $ are the input graph signals at time steps $ t-k $ and $ t $ respectively, $ Z_t^{pdf} $ denotes the updated feature at time step $ t $, $ W_{k,1}^{pdf} \in \mathbb{R}^{d \times d} $, $ W_{k,2}^{pdf} \in \mathbb{R}^{d \times d} $, and $ b^{pdf} \in \mathbb{R}^{d} $ are learnable parameters.

By this design, our STJGC simultaneously models the information propagation from three kinds of connections (i.e., spatial, temporal, and spatio-temporal) in a unified operation. 

\paragraph{\textbf{\textit{STJGC on adaptive STJG}}} As the pre-defined STJG may not reflect the underlying correlations among nodes, we further propose STJGC on adaptive STJG. The computation is similar as that on pre-defined STJG:
%
%
%
\begin{equation}
Z_t^{adt}=\sum_{k=0}^{K-1}\phi(\tilde{L}_{t-k;t}X_{t-k}W_k^{adt}+b^{adt})
\label{eq11},
\end{equation}
where $ \tilde{L}_{t-k;t} $ is the normalized adaptive STJG adjacency matrix between time steps $ t-k $ and $ t $ (defined in Equation~\ref{eq6}). Inspired by the bi-directional RNN~\cite{Schuster-and-Paliwal:TSP1997}, we consider both time directions of the information flow. Specifically, we compute two adaptive STJG adjacency matrices: $ \tilde{L}_{t-k;t} $ and $ \tilde{L}_{t;t-k} $, and modify Equation~\ref{eq11} accordingly, as:
%
%
%
\begin{equation}
Z_t^{adt}=\sum_{k=0}^{K-1}\phi(\tilde{L}_{t-k;t}X_{t-k}W_{k,1}^{adt}+\tilde{L}_{t;t-k}X_{t-k}W_{k,2}^{adt}+b^{adt}),
\label{eq12}
\end{equation}
%
%
%
where $ Z_t^{adt} $ is the updated feature at time step $ t $, which encodes the comprehensive and dynamic spatio-temporal correlations, $ W_{k,1}^{adt} \in \mathbb{R}^{d \times d} $, $ W_{k,2}^{adt} \in \mathbb{R}^{d \times d} $, and $ b^{pdf} \in \mathbb{R}^{d} $ are learnable parameters.
 
\paragraph{\textbf{\textit{Gating fusion}}} \label{Gating fusion}

The pre-defined and adaptive STJGs represent the spatio-temporal correlations from distinct perspectives. To enhance the representation ability, we use a gating mechanism to fuse the features extracted on two types of STJG. Specifically, we define a gate to control the importance of two features as:
\begin{equation}
G=sigmoid(W^g[Z_t^{pdf},Z_t^{adt}]+b^g)
\label{eq13},
\end{equation}
where $ [\cdot,\cdot] $ denotes the concatenation operation, the sigmoid function is used to control the output lies in range $ [0,1] $, $ W^g \in \mathbb{R}^{2d \times d} $ and $ b^g \in \mathbb{R}^{d} $ are learnable parameters. The gate $ G \in \mathbb{R}^{N \times d} $ controls the information flow between pre-defined and adaptive STJGs in both node-wise and channel-wise. Based on the gate, we fuse two features as: 
\begin{equation}
Z_t=G \odot Z_t^{pdf} + (1-G) \odot Z_t^{adt}
\label{eq14},
\end{equation}
where $ \odot $ denotes the element-wise product. As a result, $ Z_t \in \mathbb{R}^{N \times d} $ represents the updated representation of $ N $ nodes at time step $ t $, which aggregates the information from their spatial, temporal, and spatio-temporal neighborhoods on both types of STJG.

\subsubsection{Dilated Causal STJGC Layers} \label{Dilated Causal STJGC Layers}

\begin{figure*}
	\centering
	\includegraphics[width = 0.9 \textwidth]{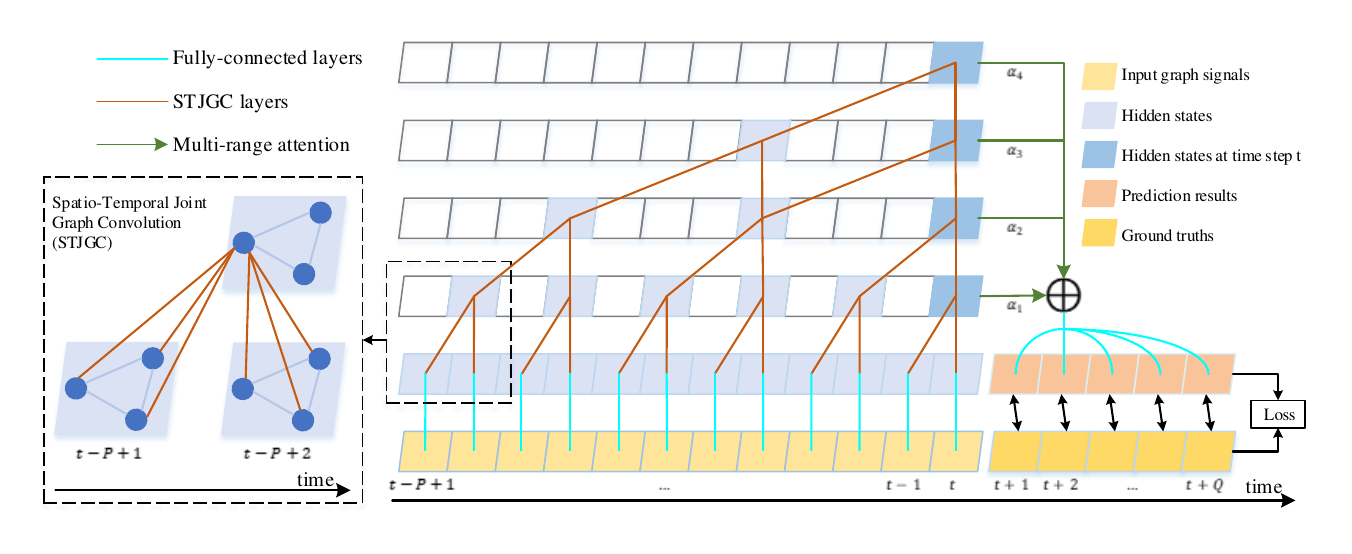}
	\caption{The illustration of the \textit{dilated causal STJGC module} (middle part in the figure) and the \textit{prediction module} (right part in the figure) in STJGCN. In the dilated csusal STJGC module, the inputs are first transformed by fully-connected layers and then be passed to the dilated causal STJGC layers, which pick inputs every $ \gamma $ (dilation factor, $ \gamma=\{1,2,4,4\} $ for each STJGC layer in the figure) step and apply STJGC (left part in the figure) to the selected inputs. The prediction module first aggregates the outputs of each STJGC layer via the multi-range attention mechanism and then uses fully-connected layers to produce the prediction results.}
	\label{fig3}
\end{figure*}

The STJGC operation is able to model the correlations in different time ranges by controlling the time difference $ k $. In addition, different STJGC layers aggregate information within diverse neighborhood ranges. This makes it flexible to model the spatio-temporal correlations in multiple neighborhood and time ranges. The information in different ranges reveals distinct traffic properties. A small range uncovers the local dependency and a large range indicates the global dependency. Inspired by the dilated causal convolution~\cite{Oord-et-al:arXiv2016,Bai-et-al:arXiv2018}, which is able to capture diverse time-ranges of dependencies in different layers, we propose dilated causal STJGC layers to capture multiple ranges of spatio-temporal dependencies. 

\paragraph{\textit{Dilated causal convolution}} 

The dilated causal convolution operation slides over the input sequence by skipping elements with a certain time step (i.e., dilation factor $ \gamma $), and it involves only historical information at each time step to satisfy the causal constraint. In this way, it models diverse time-ranges of dependencies in different layers. 

\paragraph{\textit{\textbf{Dilated causal STJGC}}} 

As illustrated in Figure~\ref{fig3}, we first transform the inputs into $ d $ dimension space using fully-connected layers. Then we stack a couple of STJGC layers upon it in the dilated causal way. Different to the standard dilated causal convolution using 1D CNN, we use the STJGC in each layer to model the dynamic and comprehensive spatio-temporal correlations. Suppose the length of input graph signals is $ P = 12 $, we could stack four STJGC layers with kernel size $ K = 2 $ and dilation factor $ \gamma = \{2, 4, 4, 4\} $ in each layer, respectively. The residual connections~\cite{He-et-al:CVPR2016} are also applied in each STJGC layer at the corresponding output time steps. The number of STJGC layers, dilation factors and kernel size could be re-designed according to the length of input graph signals, in order to ensure that the output of the last STJGC layer covers the information from all input time steps. 

In these dilated causal STJGC layers, each STJGC layer captures different ranges of spatio-temporal dependencies. For example, as shown in Figure~\ref{fig3}, in the first STJGC layer, the hidden state at time step $ t $ aggregates information from 1-hop neighborhoods at time steps $ t - 1 $ and $ t $. With the layer goes deeper, it could extract features from higher order neighborhoods at longer time-ranges. In particular, in the last STJGC layer, each node at time step $ t $ captures the information within 4-hop neighborhoods from total $ P $ time steps. 


\subsection{Prediction Module} \label{Prediction Module}

In this module, we first propose a multi-range attention mechanism to aggregate the information of different ranges extracted by the dilated causal STJGC layers, and then apply independent fully-connected layers to produce the multi-step ahead prediction results.

\subsubsection{Multi-Range Attention} 

As introduced in section~\ref{Dilated Causal STJGC Layers}, each STJGC layer captures different spatio-temporal ranges of dependencies. A small range uncovers the local dependency and a large range indicates the global dependency, e.g., the correlations between distant nodes at distant time steps. Thus, It is essential to combine the multi-range information. In addition, the importance of different ranges could be diverse. We propose a multi-range attention mechanism to aggregate the information of different ranges. Mathematically, we denote the hidden state of node $ v_i $ at time step $ t $ in $ m $-th STJGC layer as $ z_i^{(m)} \in \mathbb{R}^{d} $, the attention score is computed as:
\begin{equation}
s_i^{m}=\mathrm{v}^\top tanh(W^az_i^{(m)}+b^a)
\label{eq15},
\end{equation}
\begin{equation}
\alpha_i^m=\frac{\exp(s_i^m)}{\sum_{m=1}^{M}\exp(s_i^{m})}
\label{eq16},
\end{equation}
where $ W^a \in \mathbb{R}^{d \times d} $, $ b^a \in \mathbb{R}^{d} $, and $ \mathrm{v} \in \mathbb{R}^{d} $ are learnable parameters, $ M $ is the number of STJGC layers, and $ \alpha_i^m $ is the attention score, indicating the importance of $ z_i^{(m)} $. Based on the attention scores, the multi-range information can be aggregated as:
\begin{equation}
y_i=\sum\nolimits_{m=1}^{M}\alpha_i^mz_i^{(m)}
\label{eq17},
\end{equation}
where $ y_i $ is the updated feature of node $ v_i $, which aggregates the information from multiple spatio-temporal ranges. The attention mechanism is conducted on all of the $ N $ nodes in parallel with shared learnable parameters, and produces an output as $ Y \in \mathbb{R}^{N \times d} $. 

\subsubsection{Independent Fully-Connected Layers} 

As the traffic of different time steps may exhibit different properties, it would be better to use different networks to generate the predictions at different forecasting horizons. We thus apply $ Q $ independent two-fully-connected layers upon $ Y $ to produce the $ Q $ time steps ahead prediction results:
\begin{equation}
\hat{X}_{t+i} = \phi(YW_1^i+b_1^i)W_2^i+b_2^i
\label{eq18},
\end{equation}
where $ \hat{X}_{t+i} $ denotes the prediction result of time step $ t + i $ ($ i=1,2,\cdots,Q $), $ W_1^i \in \mathbb{R}^{d \times d} $, $ b_1^i \in \mathbb{R}^{d} $, $ W_2^i \in \mathbb{R}^{d \times 1} $, and $ b_2^i \in \mathbb{R} $ are the corresponding learnable parameters, $ \phi(\cdot) $ is an activation function.

\subsubsection{Loss Function} 

The mean absolute error (MAE) loss is commonly used in the traffic forecasting problem~\cite{Li-et-al:ICLR2018,Wu-et-al:IJCAI2019,Zheng-et-al:AAAI2020}. In practice, the MAE loss optimizes all prediction values equally regardless of the value size, which leads to relatively non-ideal predictions for small values compared to the predictions of large values. The mean absolute percentage error (MAPE) loss is more relevant to the predictions of small values. Thus, we propose to combine the MAE loss and MAPE loss as our loss function:  
%
\begin{equation}
\mathcal{L}(\hat{X}_{t+i};\Theta) = \frac{1}{Q}(\sum_{i=1}^{Q}(|\hat{X}_{t+i}-X_{t+i}| + \beta \cdot \frac{|\hat{X}_{t+i}-X_{t+i}|}{X_{t+i}} \cdot 100))
\label{eq19},
\end{equation}
where $ \beta $ is used to balance MAE loss and MAPE loss, $ \Theta $ denotes all learnable parameters in STJGCN. 


\subsection{Complexity Analysis}


We further analyze the time complexity of the main components in each module in our STJGCN.

In the \textit{STJG construction module}, the computation mainly comes from the learning of adaptive STJG adjacency matrix (Equation~\ref{eq6}). The time complexity is $ O(Nd^2+N^2d) $, where $ N $ denotes the number of nodes, $ d $ is the dimension of the spatio-temporal embedding. Regarding $ d $ as a constant, the time complexity turns to $ O(N^2) $, which is attributed to the pairwise computation of the $ N $ nodes' embeddings. 
One concern is that the large-scaled node would result in a more expensive cost. To mitigate the scale problem, we suggest to only calculate the connected edges in adaptive STJG adjacency matrix according to a priori knowledge (i.e., pre-defined STJG).

In the \textit{dilated casual STJGC module}, the time complexity mainly depends on the computation of each STJGC operation (Equations~\ref{eq10} and~\ref{eq12}), which incurs $ O(K(|\mathcal{E}|d+Nd^2)) $ time complexity. Here, $ K $ is the kernel size, $ |\mathcal{E}| $ denotes the number of edges in the graph, and $ d $ is the dimension of hidden states. The time complexity of STJGC mainly depends on $ |\mathcal{E}| $, as each node aggregates information from its neighborhoods, whose number is equal to the edge number. 

In the \textit{prediction module}, the time complexities of multi-range attention mechanism (Equations~\ref{eq15},~\ref{eq16}, and~\ref{eq17}) and independent fully-connected layers (Equation~\ref{eq18}) are $ O(N(Md+d^2)) $ and $ O(QNd^2) $, respectively. Thus, the total time complexity of the prediction module is $ O(N(Md+Qd^2)) $, where $ M $ is the number of STJGC layers and $ Q $ is the number of time steps to be predicted. The time complexity is highly related to $ Q $, as we use $ Q $ independent fully-connected layers to produce the multi-step prediction results.   

\section{Experiments} \label{Experiments}

\subsection{Datasets}

\begin{table}
	\caption{Summary statistics of five datasets.}
	\label{table2}
	\centering
	\resizebox{1.00 \columnwidth}{!}{
		\begin{tabular}{lccc}
			\toprule
			Dataset	& Time range 				& Time interval	& \# Nodes	\\
			\midrule
			PeMSD3	& 1/Sep/2018 - 30/Nov/2018	& 5-minute		& 358		\\
			PeMSD4	& 1/Jan/2018 - 28/Feb/2018	& 5-minute		& 307 		\\
			PeMSD7	& 1/May/2017 - 31/Aug/2017	& 5-minute		& 883		\\
			PeMSD8	& 1/Jul/2016 - 31/Aug/2016	& 5-minute		& 170		\\
            Seattle-Loop	& 1/Jan/2015 - 31/Dec/2015	& 5-minute		& 323		\\
			\bottomrule
		\end{tabular}}
\end{table}

We evaluate our STJGCN on five highway traffic datasets: PeMSD3, PeMSD4, PeMSD7, PeMSD8 and Seattle-Loop. The previous four datasets are released in~\cite{Guo-et-al:AAAI2019,Song-et-al:AAAI2020}. These datasets are collected by the Caltrans Performance Measurement System (PeMS) from 4 districts in real time every 30 seconds. The raw traffic data is aggregated into 5-minute time interval. There are three kinds of traffic measurements in PeMSD4 and PeMSD8 datasets, including total flow, average speed, and average occupancy. In PeMSD3 and PeMSD7 datasets, only the traffic flow is recorded. Seattle-Loop is released in~\cite{cui2018deep,cui2019traffic}, which is a highway speed dataset collected from 323 loop detectors in the Greater Seattle Area. The dataset contains 5-minute resolution traffic speed data for the entirety of 2015. Following previous studies~\cite{Bai-et-al:NIPS2020,Chen-et-al:ICML2021,shin2022pgcn}, we predict the traffic flow in first four datasets, and traffic speed in last dataset. The summary statistics of five datasets are presented in Table~\ref{table2}.





All datasets are normalized using the Z-Score method, and be split in chronological order with 60\% for training, 20\% for validation, and 20\% for testing. The pair-wise road network distances are provided in the datasets, and we use them to construct the pre-defined STJG according to Equation~\ref{eq3}. 




\subsection{Experimental Setup} \label{section4.2}

\subsubsection{Evaluation Metrics}

We adopt three widely used metrics for evaluation, i.e., mean absolute error (MAE), root mean squared error (RMSE), and mean absolute percentage error (MAPE), which are defined as: 
\begin{equation}
MAE = \frac{1}{NQ}\sum_{i=1}^{N}\sum_{j=1}^{Q}|\hat{X}_{i,t+j}-X_{i,t+j}|
\label{eq20},
\end{equation}
\begin{equation}
RMSE = \sqrt{\frac{1}{NQ}\sum_{i=1}^{N}\sum_{j=1}^{Q}(\hat{X}_{i,t+j}-X_{i,t+j})^2}
\label{eq21},
\end{equation}
\begin{equation}
MAPE = \frac{1}{NQ}\sum_{i=1}^{N}\sum_{j=1}^{Q}\frac{|\hat{X}_{i,t+j}-X_{i,t+j}|}{X_{i,t+j}}
\label{eq22},
\end{equation}
where $ \hat{X}_{i,t+j} $ and $ X_{i,t+j} $ denote the prediction result and ground truth of node $ v_i $ at time step $ t + j $, respectively, $ N $ is the number of nodes, and $ Q $ is the number of time steps to be predicted.

\subsubsection{Experimental Settings} 

\begin{table}
	\centering
	\caption{Hyperparameter settings of STJGCN on five datasets.}
	\begin{tabular}{lccccc}
		\toprule
		Dataset	& $ \delta_{pdf} $	& $ \delta_{adt} $ 	& $ d $	& $ K $	& $ \beta $	\\
		\midrule
		PeMSD3	& 0.5				& 0.5				& 64	& 2 	& 0.1		\\
		PeMSD4	& 0.5				& 0.5				& 64	& 3 	& 1.0		\\
		PeMSD7	& 0.9				& 0.7				& 64	& 2		& 0.5		\\
		PeMSD8	& 0.5				& 0.3				& 64	& 2		& 1.5		\\
        Seattle-Loop	& 0.5				& 0.3				& 64	& 2		& 0.1		\\
		\bottomrule
	\end{tabular}
	\label{table3}
\end{table}

The PeMSD3 and PeMSD7 datasets contain one traffic measurement (i.e., traffic flow). Thus, the dimensions of the input and output are $ C=1 $ and 1, respectively. The PeMSD4 and PeMSD8 datasets contain three traffic measurements (i.e., traffic flow, average speed, and average occupancy), and only the traffic flow is predicted in the experiments~\cite{Bai-et-al:NIPS2020,Chen-et-al:ICML2021}. Thus, the dimensions of the input and output are $ C=3 $ and 1, respectively. The Seattle-Loop dataset contains one traffic measurement (i.e., traffic speed). Thus, the dimensions of the input and output are $ C=1 $ and 1. Following previous studies~\cite{Bai-et-al:NIPS2020,Chen-et-al:ICML2021,shin2022pgcn}, we use the traffic data of historical 12 time steps ($ P = 12 $) to forecast the next 12 time steps ($ Q = 12 $). 

The core hyperparameters in STJGCN include the thresholds $ \delta_{pdf} $ and $ \delta_{adt} $ in pre-defined and adaptive STJG adjacency matrices respectively, the dimension $ d $ of hidden states, the kernel size $ K $ of each STJGC layer, and the threshold $ \beta $ in the loss function. We tune these hyperparameters on the validation set that achieve the best validation performance. We provide a parameter study in section~\ref{Parameter Study}. The detailed hyperparameter settings of STJGCN on five datasets are presented in Table~\ref{table3}.  

The nonlinear activation function $ \phi(\cdot) $ in our STJGCN refers to the ReLU activation~\cite{Nair-and-Hinton:ICML2010}, and we also add a Batch Normalization~\cite{Ioffe-and-Szegedy:ICML2015} layer before each ReLU activation function. 

We train our model using the Adam optimizer~\cite{Kingma-and-Ba:ICLR2015} with an initial learning rate 0.001 and batch size 64 on a NVIDIA Tesla V100 GPU card. We run the experiments for 200 epochs and save the best model that evaluated on the validation set. We run each experiment 5 times, and report the mean errors and standard deviations.

\subsubsection{Baseline Methods}

We compare STJGCN with 11 baseline methods, which could be divided into two categories. The first category is the time-series prediction models, including:

\begin{itemize}
	\item VAR~\cite{VAR:1994}: Vector Auto-Regressive is a traditional time-series model, which can capture pairwise relationships among all traffic series.
	\item FC-LSTM~\cite{Sutskever-et-al:NIPS2014}: an encoder-decoder framework using long short-term memory (LSTM) with peephole for multi-step time-series prediction.
	\item SVR~\cite{Drucker-et-al:NIPS1997}: Support Vector Regression utilizes a linear support vector machine to perform regression.
\end{itemize}

The second category refers to the spatio-temporal graph neural networks, which are detailed as follows:

\begin{itemize}
	\item DCRNN~\cite{Li-et-al:ICLR2018}: Diffusion Convolutional Recurrent Neural Network, which models the traffic as a diffusion process, and integrates diffusion convolution with recurrent neural network (RNN) into the encoder-decoder architecture. 
	\item STGCN~\cite{Yu-et-al:IJCAI2018}: Spatio-Temporal Graph Convolutional Network, which employs graph convolutional network (GCN) to capture spatial dependencies and 1D convolutional neural network (CNN) for temporal correlations modeling.
	\item ASTGCN~\cite{Guo-et-al:AAAI2019}: Attention based Spatio-Temporal Graph Convolutional Network that designs spatial and temporal attention mechanisms to capture spatial and temporal patterns, respectively. 
	\item Graph WaveNet~\cite{Wu-et-al:IJCAI2019}: a graph neural network that performs diffusion convolution with both pre-defined and self-adaptive adjacency matrices to capture spatial dependencies, and applies 1D dilated causal convolution to capture temporal dependencies.
	\item STSGCN~\cite{Song-et-al:AAAI2020}: Spatio-Temporal Synchronous Graph Convolutional Network that designs spatio-temporal synchronous modeling mechanism to capture localized spatio-temporal correlations.
	\item AGCRN~\cite{Bai-et-al:NIPS2020}: Adaptive Graph Convolutional Recurrent Network that learns data-adaptive adjacency matrix for graph convolution to model spatial correlations and uses gated recurrent unit (GRU) to model temporal correlations.   
	\item GMAN~\cite{Zheng-et-al:AAAI2020}: Graph Multi-Attention Network is an encoder-decoder framework, which designs multiple spatial and temporal attention mechanisms in the encoder and decoder to model spatio-temporal correlations, and a transform attention mechanism to transform information from encoder to decoder.
	\item Z-GCNETs~\cite{Chen-et-al:ICML2021}: Time Zigzags at Graph Convolutional Networks that introduce the concept of zigzag persistence~\cite{Carlsson-and-Silva:2010} into the graph convolutional networks for modeling the spatial correlations and use the GRU networks to capture the temporal dependencies.
\end{itemize}

\begin{table*}
	\caption{Forecasting performance comparison of different models on five datasets.}
	\label{table4}
	\centering
	\resizebox{1.00 \textwidth}{!}{
	\begin{tabular}{llccccccccccc|c}
		\toprule
		Dataset					&Metrics   &VAR   &SVR   &FC-LSTM 		 &DCRNN          &STGCN          &ASTGCN         &Graph WaveNet  &STSGCN 		 &AGCRN  		 &GMAN			 &Z-GCNETs       &STJGCN \\
		\midrule
		\multirow{3}{*}{PeMSD3} &MAE	   &19.72 &19.77 &19.56$\pm$0.32 &17.62$\pm$0.13 &19.76$\pm$0.67 &18.67$\pm$0.42 &15.67$\pm$0.06 &15.74$\pm$0.09 &16.10$\pm$0.16 &15.52$\pm$0.09 &15.90$\pm$0.77 &\textbf{14.92$\pm$0.10} \\
								&RMSE      &32.38 &32.78 &33.38$\pm$0.46 &29.86$\pm$0.47 &33.87$\pm$1.18 &30.71$\pm$1.02 &26.42$\pm$0.14 &26.39$\pm$0.36 &28.55$\pm$0.28 &26.53$\pm$0.19 &27.90$\pm$0.86 &\textbf{25.70$\pm$0.41} \\
								&MAPE (\%) &20.50 &23.04 &19.56$\pm$0.51 &16.83$\pm$0.13 &17.33$\pm$0.94 &19.85$\pm$1.06 &15.72$\pm$0.23 &15.40$\pm$0.07 &15.02$\pm$0.26 &15.19$\pm$0.25 &15.51$\pm$1.67 &\textbf{14.81$\pm$0.16} \\
		\midrule
		\multirow{3}{*}{PeMSD4} &MAE       &24.44 &26.18 &23.60$\pm$0.52 &24.42$\pm$0.06 &23.90$\pm$0.17 &22.90$\pm$0.20 &19.91$\pm$0.10 &19.62$\pm$0.16 &19.74$\pm$0.09 &19.25$\pm$0.06 &19.54$\pm$0.07 &\textbf{18.81$\pm$0.06} \\
								&RMSE	   &37.76 &38.91 &37.11$\pm$0.50 &37.48$\pm$0.10 &36.43$\pm$0.22 &35.59$\pm$0.35 &31.06$\pm$0.17 &31.02$\pm$0.29 &32.01$\pm$0.17 &30.85$\pm$0.21 &31.33$\pm$0.11 &\textbf{30.35$\pm$0.09} \\
								&MAPE (\%) &17.27 &22.84 &16.17$\pm$0.13 &16.86$\pm$0.09 &13.67$\pm$0.14 &16.75$\pm$0.59 &13.62$\pm$0.22 &13.13$\pm$0.11 &12.98$\pm$0.21 &13.00$\pm$0.26 &12.87$\pm$0.05 &\textbf{11.92$\pm$0.04} \\
		\midrule
		\multirow{3}{*}{PeMSD7} &MAE       &27.96 &28.45 &34.05$\pm$0.51 &24.45$\pm$0.85 &26.22$\pm$0.37 &28.13$\pm$0.70 &20.83$\pm$0.18 &21.64$\pm$0.11 &21.22$\pm$0.17 &20.68$\pm$0.08 &21.26$\pm$0.28 &\textbf{19.95$\pm$0.04} \\
								&RMSE      &41.31 &42.67 &55.70$\pm$0.60 &37.61$\pm$1.18 &39.18$\pm$0.42 &43.67$\pm$1.33 &33.64$\pm$0.22 &34.87$\pm$0.27 &35.05$\pm$0.13 &33.56$\pm$0.12 &34.53$\pm$0.28 &\textbf{33.01$\pm$0.07} \\
								&MAPE (\%) &12.11 &14.00 &15.31$\pm$0.31 &10.67$\pm$0.53 &10.74$\pm$0.16 &13.31$\pm$0.55 &9.10$\pm$0.27  &9.09$\pm$0.05  &9.00$\pm$0.12  &9.31$\pm$0.12  &9.04$\pm$0.11  &\textbf{8.31$\pm$0.11} \\
		\midrule
		\multirow{3}{*}{PeMSD8} &MAE       &19.83 &20.92 &21.18$\pm$0.27 &18.49$\pm$0.16 &18.79$\pm$0.49 &18.72$\pm$0.16 &15.57$\pm$0.12 &16.12$\pm$0.25 &15.92$\pm$0.19 &14.87$\pm$0.15 &16.12$\pm$0.08 &\textbf{14.53$\pm$0.17} \\
								&RMSE      &29.24 &31.23 &31.88$\pm$0.43 &27.30$\pm$0.22 &28.23$\pm$0.36 &28.99$\pm$0.11 &24.32$\pm$0.21 &24.89$\pm$0.52 &25.31$\pm$0.25 &24.06$\pm$0.16 &25.74$\pm$0.13 &\textbf{23.74$\pm$0.20} \\
								&MAPE (\%) &13.08 &14.24 &13.72$\pm$0.27 &11.69$\pm$0.06 &10.55$\pm$0.30 &12.53$\pm$0.48 &10.32$\pm$0.79 &10.50$\pm$0.22 &10.30$\pm$0.13 &9.77$\pm$0.07  &10.35$\pm$0.09 &\textbf{9.15$\pm$0.09} \\
        \midrule
		\multirow{3}{*}{Seattle-Loop} & MAE       &3.77 &4.86 &3.94$\pm$0.06 &3.54$\pm$0.04 &3.55$\pm$0.09 &3.37$\pm$0.04 &3.81$\pm$0.03 &3.52$\pm$0.05 &3.33$\pm$0.04 &3.22$\pm$0.03 &3.29$\pm$0.02 &\textbf{3.19$\pm$0.03} \\
								&RMSE      &5.86 &8.96 &7.42$\pm$0.09 &6.22$\pm$0.04 &5.95$\pm$0.07 &5.69$\pm$0.03 &6.81$\pm$0.04 &6.32$\pm$0.11 &5.99$\pm$0.05 &5.70$\pm$0.03 &5.85$\pm$0.02 &\textbf{5.61$\pm$0.04} \\
								&MAPE  (\%) &11.12 &15.38 &11.74$\pm$0.05 &10.63$\pm$0.01 &9.43$\pm$0.06 &10.14$\pm$0.09 &10.73$\pm$0.14 &10.14$\pm$0.04 &9.74$\pm$0.03 &9.27$\pm$0.01  &9.40$\pm$0.02 &\textbf{8.92$\pm$0.02} \\
		\bottomrule
	\end{tabular}}
\end{table*}

\subsection{Experimental Results}

\subsubsection{Overall Comparison}

Table~\ref{table4} presents the forecasting performance comparison of our STJGCN with 11 baseline methods. We observe that: (1) the time-series prediction models, including traditional approach (i.e., VAR), machine learning based method (i.e., SVR), and deep neural network (i.e., FC-LSTM) perform poorly as they only consider the temporal correlations. (2) Spatio-temporal graph neural networks generally achieve better performances as they further model the spatial correlations using graph neural networks. (3) Our STJGCN performs the best in terms of all metrics on all datasets (1.4\%\textasciitilde7.7\% improvement against the second best results). Compared with other graph-based methods, the advantages of our STJGCN are three-fold. First, STJGCN models comprehensive spatio-temporal correlations. Second, STJGCN is able to capture dynamic dependencies at different time steps. Third, STJGCN leverages the information of multiple spatio-temporal ranges. 

\subsubsection{Ablation Study}

\begin{table*}[htb]
	\caption{Effect of spatio-temporal connections, dynamic graph modeling, multi-range information, and independent fully-connected layers.}
	\label{table5}
	\centering
	\resizebox{1.00 \textwidth}{!}{
		\begin{tabular}{llcccccccc}
			\toprule
			Dataset					& Metrics	& STJGCN				  & w/o STC-pdf	   & w/o STC-adt	& w/o STC		 & w/o dgm		  & w/o mr		   & w/o att		& w/o idp 		 \\
			\midrule
			\multirow{3}{*}{PeMSD4} & MAE       & \textbf{18.81$\pm$0.06} & 18.99$\pm$0.14 & 19.07$\pm$0.10	& 19.36$\pm$0.09 & 19.70$\pm$0.06 & 19.03$\pm$0.04 & 18.97$\pm$0.09 & 18.89$\pm$0.08 \\
									& RMSE  	& \textbf{30.35$\pm$0.09} & 30.63$\pm$0.23 & 30.71$\pm$0.13	& 30.80$\pm$0.10 & 31.47$\pm$0.05 & 30.79$\pm$0.08 & 30.56$\pm$0.12 & 30.46$\pm$0.10 \\
									& MAPE (\%)	& \textbf{11.92$\pm$0.04} & 12.00$\pm$0.07 & 12.07$\pm$0.06	& 12.27$\pm$0.08 & 12.39$\pm$0.07 & 11.98$\pm$0.03 & 11.96$\pm$0.02 & 11.95$\pm$0.02 \\
			\midrule
			\multirow{3}{*}{PeMSD8} & MAE       & \textbf{14.53$\pm$0.17} & 14.63$\pm$0.23 & 14.82$\pm$0.09	& 15.07$\pm$0.07 & 15.49$\pm$0.22 & 15.11$\pm$0.57 & 14.67$\pm$0.11	& 14.60$\pm$0.11 \\
									& RMSE  	& \textbf{23.74$\pm$0.20} & 24.01$\pm$0.22 & 24.11$\pm$0.14	& 24.22$\pm$0.14 & 24.49$\pm$0.23 & 24.49$\pm$0.55 & 24.03$\pm$0.30	& 23.96$\pm$0.21 \\
									& MAPE (\%)	& \textbf{9.15$\pm$0.09}  & 9.18$\pm$0.19  &  9.26$\pm$0.08	&  9.48$\pm$0.06 &  9.55$\pm$0.16 &  9.39$\pm$0.22 & 9.16$\pm$0.09	& 9.16$\pm$0.12  \\	
			\bottomrule
	\end{tabular}}
\end{table*}

To better understand the effectiveness of different components in STJGCN, we conduct ablation studies on PeMSD4 and PeMSD8 datasets.  

\paragraph{\textit{Effect of spatio-temporal connections}} 

One difference between our STJG with normal spatio-temporal graph is that we explicitly add the spatio-temporal connections between different nodes at different time steps. To evaluate the effectiveness of this approach, we drop them separately/simultaneously from the pre-defined or/and adaptive STJG. These three variants of STJGCN are named as ``w/o STC-pdf'' (drop in pre-defined STJG), ``w/o STC-adt'' (drop in adaptive STJG), and ``w/o STC'' (drop in both types of STJG), respectively. The results in Table~\ref{table5} demonstrate that the introduction of spatio-temporal connections improves the performance as it helps the model to explicitly capture comprehensive spatio-temporal correlations. 

\paragraph{\textit{Effect of dynamic graph modeling}} 

To evaluate the effect of dynamic graph modeling, we conduct experiments of learning static adjacency matrices. Specifically, we design a variant of STJGCN (i.e., ``w/o dgm'') that only uses the node embedding to generate the adaptive STJG adjacency matrix without using the time feature. The results in Table~\ref{table5} validate the effectiveness of modeling dynamic correlations among nodes at different time steps. 

\paragraph{\textit{Effect of multi-range information}} 

To verify the effect of multi-range information, we design a variant of STJGCN, namely ``w/o mr'', in which we do not combine multiple ranges of information but directly use the output of the last STJGC layer to produce the predictions. The results in Table~\ref{table5} indicate the necessity of leveraging multi-range information. We further design a variant ``w/o att'' that directly adds the outputs of each STJGC layer together without using the multi-range attention mechanism, and it performs worse than STJGCN, showing that it is beneficial to distinguish the importance of different ranges of information.  

\paragraph{\textit{Effect of independent fully-connected layers}} 

In the prediction module, we use $ Q $ independent fully-connected layers to produce the multi-step predictions. To evaluate the effectiveness of this, we conduct experiments of using shared fully-connected layers with $ Q $ units in the output layer to produce the $ Q $ time steps predictions. We name this variant of STJGCN as ``w/o idp'', and present the experimental results in Table~\ref{table5}.  We observe that STJGCN improves the performances by introducing independent learning parameters for multi-step prediction. A potential reason is that the traffic of different time steps may exhibit different properties, and using different networks to generate the predictions at different forecasting horizons could be beneficial. 

\paragraph{\textit{Effect of different STJG adjacency matrix configurations}}

We further conduct experiments of using different STJG adjacency matrix configurations to evaluate their effectiveness. As shown in Table~\ref{table6}, the models with only pre-defined STJG adjacency matrices (lines 3-4) achieve poor performances as they do not capture the underlying dependencies in the data. We observe that the models with only adaptive STJG adjacency matrices (lines 5-6) could realize promising performances, which indicates that our model can also be used even if the graph structure is unavailable. By using both pre-defined and adaptive STJG adjacency matrices (line 7), we could achieve better results. We further apply a gating fusion approach (section~\ref{Gating fusion}) in STJGCN (line 8) and observe consistent improvement of the predictive performances, as the gate is able to control the information flow between pre-defined and adaptive STJGs.    

\begin{table*}[htb]
	\caption{Effect of different STJG adjacency matrix configurations. The term ``gf'' in the last line denotes the gating fusion approach.}
	\label{table6}
	\centering
	\begin{tabular}{lcccccr}
		\toprule
		\multirow{2}{*}{STJG adjacency matrix configuration}										& \multicolumn{3}{c}{PeMSD4}														& \multicolumn{3}{c}{PeMSD8}														\\
		\cmidrule(r){2-4}\cmidrule(r){5-7}
		& MAE						& RMSE						& MAPE (\%)					& MAE						& RMSE						& MAPE (\%)					\\
		\midrule	
		$ [ A_{fw}^{(k)} ] $																& 24.64$\pm$0.05			& 38.21$\pm$0.02			& 15.70$\pm$0.08			& 18.52$\pm$0.10			& 29.24$\pm$0.18			& 11.35$\pm$0.08			\\
		$ [ A_{fw}^{(k)},A_{bw}^{(k)} ] $													& 24.40$\pm$0.06			& 38.03$\pm$0.23			& 15.47$\pm$0.03			& 18.12$\pm$0.07			& 28.49$\pm$0.16			& 11.19$\pm$0.11			\\
		$ [ \tilde{L}_{t-k;t} ] $															& 19.39$\pm$0.12			& 31.60$\pm$0.23			& 12.38$\pm$0.08			& 15.93$\pm$0.15			& 25.87$\pm$0.23			& 9.98$\pm$0.07				\\
		$ [ \tilde{L}_{t-k;t},\tilde{L}_{t;t-k} ] $     									& 19.35$\pm$0.13			& 31.47$\pm$0.16			& 12.34$\pm$0.14			& 15.42$\pm$0.15			& 24.80$\pm$0.32			& 9.85$\pm$0.14				\\
		$ [ A_{fw}^{(k)},A_{bw}^{(k)},\tilde{L}_{t-k;t},\tilde{L}_{t;t-k} ] $				& 18.93$\pm$0.09			& 30.48$\pm$0.13			& 11.97$\pm$0.04 			& 14.65$\pm$0.08			& 23.93$\pm$0.14			& 9.23$\pm$0.08				\\
		\midrule	
		$ [ A_{fw}^{(k)},A_{bw}^{(k)},\tilde{L}_{t-k;t},\tilde{L}_{t;t-k} ] $ + gf (ours)	& \textbf{18.81$\pm$0.06}	& \textbf{30.35$\pm$0.09}	& \textbf{11.92$\pm$0.04}	& \textbf{14.53$\pm$0.17}	& \textbf{23.74$\pm$0.20}	& \textbf{9.15$\pm$0.09}	\\	
		\bottomrule
	\end{tabular}
\end{table*} 

\subsubsection{Parameter Study} \label{Parameter Study}

\begin{figure*}[htb]
	\centering
	\subfigure[]{
		\begin{minipage}{0.2 \textwidth}
			\label{fig4(a)} 
			\includegraphics[width = 0.95 \textwidth]{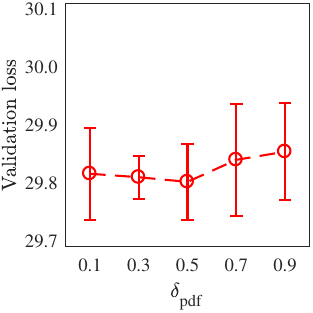}
	\end{minipage}}
	\subfigure[]{
		\begin{minipage}{0.2 \textwidth}
			\label{fig4(b)} 
			\includegraphics[width = 0.95 \textwidth]{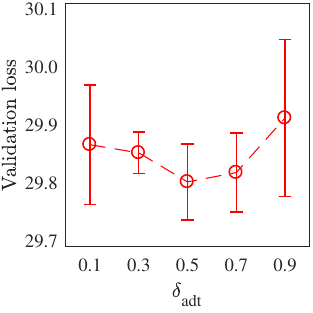}
	\end{minipage}}
	\subfigure[]{
		\begin{minipage}{0.2 \textwidth}
			\label{fig4(c)} 
			\includegraphics[width = 0.95 \textwidth]{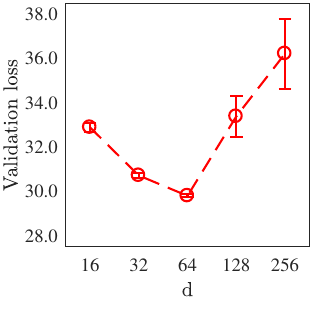}
	\end{minipage}}
	\subfigure[]{
		\begin{minipage}{0.2 \textwidth}
			\label{fig4(d)} 
			\includegraphics[width = 0.95 \textwidth]{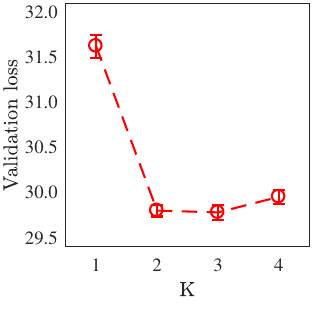}
	\end{minipage}}
	\subfigure[]{
		\begin{minipage}{0.2 \textwidth}
			\label{fig4(e)} 
			\includegraphics[width = 0.95 \textwidth]{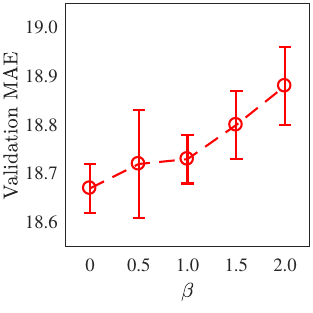}
	\end{minipage}}
	\subfigure[]{
		\begin{minipage}{0.2 \textwidth}
			\label{fig4(f)} 
			\includegraphics[width = 0.95 \textwidth]{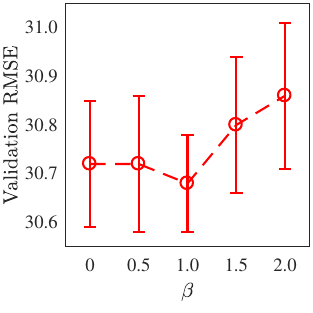}
	\end{minipage}}
	\subfigure[]{
		\begin{minipage}{0.2 \textwidth}
			\label{fig4(g)} 
			\includegraphics[width = 0.95 \textwidth]{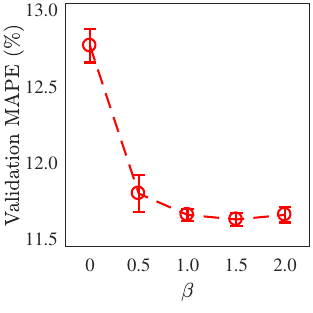}
	\end{minipage}}
	\caption{Parameter study on the PeMSD4 dataset. 
	}
	\label{fig4}
\end{figure*}

\begin{figure*}
	\centering
	\subfigure[]{
		\begin{minipage}{0.2 \textwidth}
			\label{fig5(a)} 
			\includegraphics[width = 0.95 \textwidth]{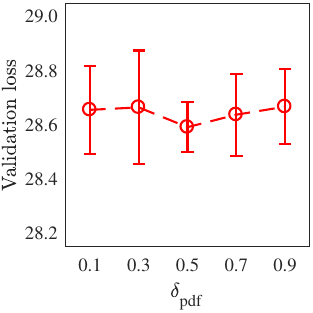}
	\end{minipage}}
	\subfigure[]{
		\begin{minipage}{0.2 \textwidth}
			\label{fig5(b)} 
			\includegraphics[width = 0.95 \textwidth]{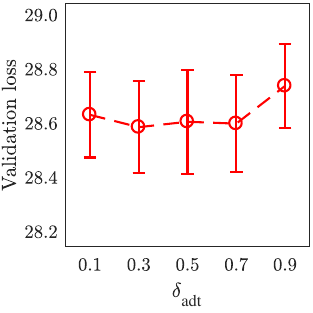}
	\end{minipage}}
	\subfigure[]{
		\begin{minipage}{0.2 \textwidth}
			\label{fig5(c)} 
			\includegraphics[width = 0.95 \textwidth]{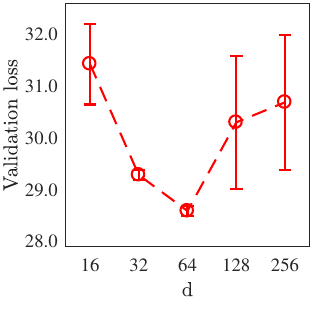}
	\end{minipage}}
	\subfigure[]{
		\begin{minipage}{0.2 \textwidth}
			\label{fig5(d)} 
			\includegraphics[width = 0.95 \textwidth]{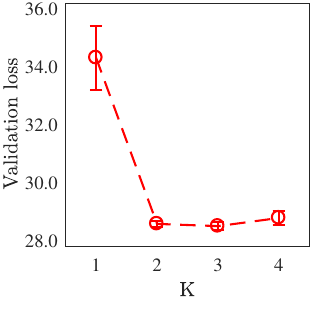}
	\end{minipage}}
	\subfigure[]{
		\begin{minipage}{0.2 \textwidth}
			\label{fig5(e)} 
			\includegraphics[width = 0.95 \textwidth]{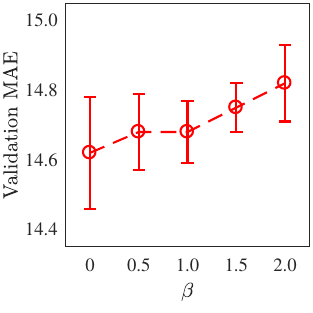}
	\end{minipage}}
	\subfigure[]{
		\begin{minipage}{0.2 \textwidth}
			\label{fig5(f)} 
			\includegraphics[width = 0.95 \textwidth]{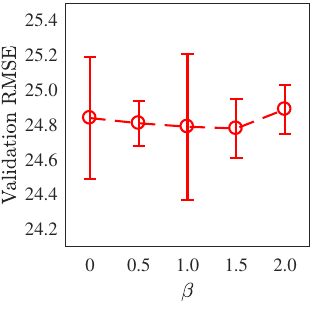}
	\end{minipage}}
	\subfigure[]{
		\begin{minipage}{0.2 \textwidth}
			\label{fig5(g)} 
			\includegraphics[width = 0.95 \textwidth]{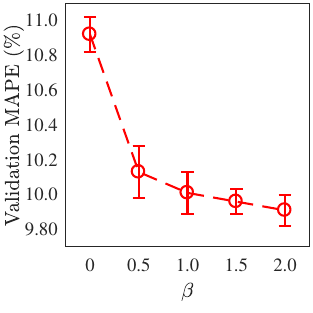}
	\end{minipage}}
	\caption{Parameter study on the PeMSD8 dataset. 
	}
	\label{fig5}
\end{figure*}

\begin{figure*}
	\centering
	\subfigure[]{
		\begin{minipage}{0.28 \textwidth}
			\label{fig6(a)} 
			\includegraphics[width = 0.95 \textwidth]{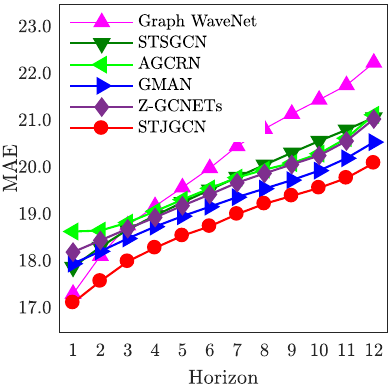}
	\end{minipage}}
	\subfigure[]{
		\begin{minipage}{0.28 \textwidth}
			\label{fig6(b)} 
			\includegraphics[width = 0.95 \textwidth]{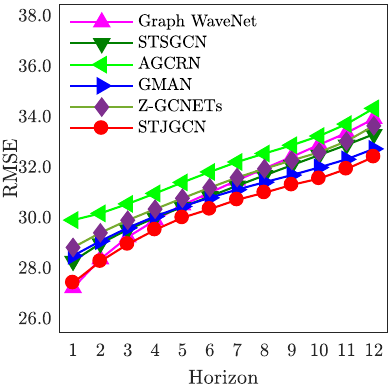}
	\end{minipage}}
	\subfigure[]{
		\begin{minipage}{0.28 \textwidth}
			\label{fig6(c)} 
			\includegraphics[width = 0.95 \textwidth]{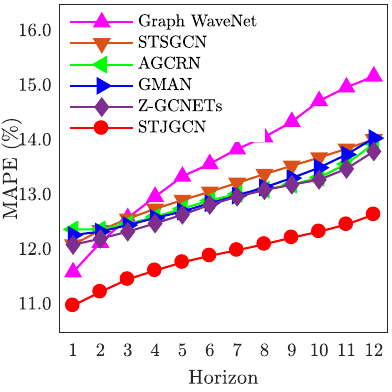}
	\end{minipage}}
	\caption{Forecasting performance comparison at each horizon on the PeMSD4 dataset.}
	\label{fig6}
\end{figure*}

\begin{figure*}
	\centering
	\subfigure[]{
		\begin{minipage}{0.28 \textwidth}
			\label{fig7(a)} 
			\includegraphics[width = 0.95 \textwidth]{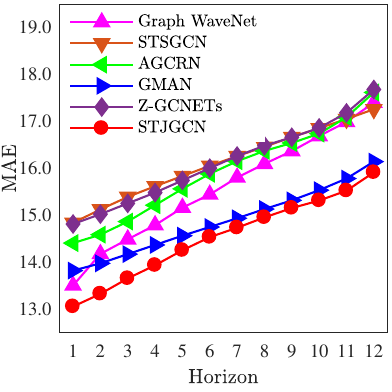}
	\end{minipage}}
	\subfigure[]{
		\begin{minipage}{0.28 \textwidth}
			\label{fig7(b)} 
			\includegraphics[width = 0.95 \textwidth]{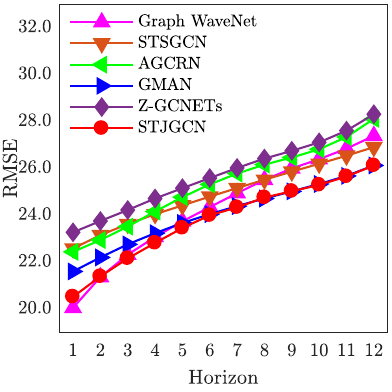}
	\end{minipage}}
	\subfigure[]{
		\begin{minipage}{0.28 \textwidth}
			\label{fig7(c)} 
			\includegraphics[width = 0.95 \textwidth]{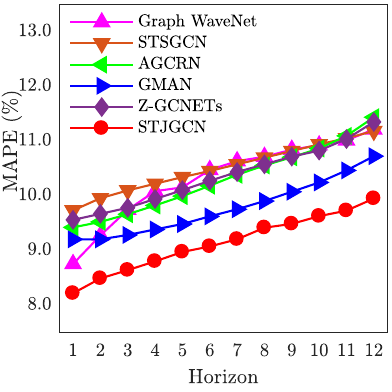}
	\end{minipage}}
	\caption{Forecasting performance comparison at each horizon on the PeMSD8 dataset.}
	\label{fig7}
\end{figure*}

\begin{table*}
	\centering
	\caption{Comparisons of parameter number and computation time. The training time is the time cost per epoch in the training phase, and the inference time is the total time cost on the validation set.}
	\resizebox{1.00 \textwidth}{!}{
	\begin{tabular}{llccccccccc}
		\toprule
		Dataset					& 							& DCRNN	& STGCN	& Graph WaveNet	& ASTGCN	& STSGCN	& AGCRN		& GMAN		& Z-GCNETs	& STJGCN	\\
		\midrule
		\multirow{3}{*}{PeMSD3} & \# Parameter (M)			& 0.37	& 0.42	& 0.31 			& 0.59		& 3.50		& 0.75 		& 0.57		& 0.52		& 0.32		\\
								& Training time (s/epoch)  	& 118.06& 12.20	& 59.73	   	 	& 78.69     & 127.86	& 55.45		& 168.77	& 208.55	& 49.82		\\
								& Inference time (s)		& 18.70	& 19.10	& 5.16	    	& 26.80		& 15.41		& 8.44		& 17.45		& 25.79		& 5.22		\\
		\midrule
		\multirow{3}{*}{PeMSD4} & \# Parameter (M)			& 0.37	& 0.38	& 0.31 			& 0.45		& 2.87		& 0.75		& 0.57		& 0.52		& 0.31		\\
								& Training time (s/epoch)  	& 69.55	& 6.54	& 32.40	   	 	& 53.51     & 56.18		& 37.05		& 82.40		& 88.41		& 25.64		\\
								& Inference time (s)		& 11.97	& 13.44	& 2.60	    	& 14.67		& 6.03		& 5.55		& 9.16		& 11.84		& 2.87		\\
		\midrule
		\multirow{3}{*}{PeMSD7} & \# Parameter (M)			& 0.37	& 0.75	& 0.31 			& 3.24		& 15.36		& 0.75 		& 0.57		& 0.52		& 0.36		\\
								& Training time (s/epoch)  	& 306.66& 33.59	& 173.85   	 	& 213.30    & 465.12	& 189.48	& 779.12	& 624.32	& 158.64	\\
								& Inference time (s)		& 45.13	& 71.17	& 16.17	    	& 64.81		& 54.60		& 26.31		& 83.2		& 89.99		& 16.30		\\
		\midrule
		\multirow{3}{*}{PeMSD8} & \# Parameter (M)			& 0.37	& 0.30	& 0.31 			& 0.18		& 1.66		& 0.75		& 0.57		& 0.52		& 0.31		\\
								& Training time (s/epoch)  	& 46.41	& 4.24	& 20.48	    	& 47.07     & 31.23		& 21.74		& 32.27		& 52.51		& 17.60		\\
								& Inference time (s)		& 8.81	& 9.37	& 1.72	    	& 14.01		& 3.09		& 3.04		& 4.06		& 7.36		& 1.67		\\	
        \midrule
		\multirow{3}{*}{Seattle-Loop} & \# Parameter (M)			& 0.37	& 0.39	& 0.31 			& 0.49		& 3.50		& 0.75		& 0.57		& 0.52		& 0.32		\\
		& Training time (s/epoch)  	& 378.16	& 59.83	& 100.82	    	& 249.34     & 120.66		& 161.44		& 1901.50		& 809.35		& 516.69		\\
		& nference time (s)		& 106.47	& 124.36	& 120.20	    	& 100.46		& 36.45		& 16.36		& 170.7		& 66.83		& 45.01		\\	
		\bottomrule
	\end{tabular}}
	\label{table7}
\end{table*}

We conduct a parameter study on five core hyperparameters in STJGCN on the PeMSD4 and PeMSD8 datasets, including the thresholds $ \delta_{pdf} $ and $ \delta_{adt} $ in the pre-defined and adaptive STJG adjacency matrices, respectively, the dimension $ d $ of hidden states, the kernel size $ K $ in the STJGC operation, and the threshold $ \beta $ in the loss function. We change the parameter under investigation and fix other parameters in each experiment. Figures~\ref{fig4} and~\ref{fig5} show the experimental results on the PeMSD4 and PeMSD8 datasets, respectively.

As shown in Figures~\ref{fig4(a)},~\ref{fig4(b)},~\ref{fig5(a)}, and~\ref{fig5(b)}, the performance is not strongly sensitive to the sparsity of the STJG adjacency matrices, which we think is because the adaptive STJG adjacency matrix could adjust itself for aggregating the neighboring information during the training stage. While, in general, a more sparse adjacency matrix is beneficial to select the most related nodes for each node, and leads to better results. However, a too sparse graph may lose the connections between interrelated nodes, and thus degrades the performances. According to the validation loss, we set $ \delta_{pdf}=\delta_{adt}=0.5 $ in the PeMSD4 dataset, and $ \delta_{pdf}=0.5 $, $ \delta_{adt}=0.3 $ in the PeMSD8 dataset.

As shown in Figures~\ref{fig4(c)} and~\ref{fig5(c)}, increasing the number of hidden units could enhance the model's expressive capacity. However, when it is larger than 64, the performance degrades significantly, as the model needs to learn more parameters and may suffer from the over-fitting problem.

Figures~\ref{fig4(d)} and~\ref{fig5(d)} show that the model performs poorly when the kernel size equals to 1, as it captures only the spatial dependencies and does not consider the correlations in the temporal dimension. We can further observe that it is enough to aggregate the information from neighboring 2 or 3 time steps at each time step. When $ K=4 $, the model's performance degrades. It is possibly because that a node's information at a time step may only correlated to the nodes at a limited number of neighboring time steps, and a large $ K $ would introduce noises into the model. Thus, according to the validation loss, we set $ K=3 $ and $ K=2 $ on the PeMSD4 and PeMSD8 datasets, respectively.

In the parameter study of the threshold $ \beta $ in the loss function, we report the validation MAE, RMSE, and MAPE instead of reporting the loss value, as the size of $ \beta $ directly impacts the size of the loss value. As shown in Figures~\ref{fig4(e)},~\ref{fig4(g)},~\ref{fig5(e)}, and~\ref{fig5(g)}, a larger $ \beta $ means the model optimizes more on the MAPE loss and less on the MAE loss, and thus leads to smaller MAPE and larger MAE. The RMSE can also be influenced, as shown in Figures~\ref{fig4(f)} and~\ref{fig5(f)}. Through a comprehensive consideration of the validation MAE, RMSE, MAPE and their standard deviations, we choose to use $ \beta=1.0 $ and $ \beta=1.5 $ in the PeMSD4 and PeMSD8 datasets, respectively.

\subsubsection{Performance Comparison at Each Horizon}

Figures~\ref{fig6} and~\ref{fig7} present the forecasting performance comparison of our STJGCN with five representative baseline methods (i.e., Graph WaveNet, STSGCN, AGCRN, GMAN, and Z-GCNETs) at each prediction time step on the PeMSD4 and PeMSD8 datasets, respectively. We exclude other baseline methods due to their poorer performances, as shown in Table~\ref{table4}. We can observe that Graph WaveNet performs well in the short-term (one or two time steps ahead) prediction. However, its performance degrades quickly with the increase of the forecasting horizon. The performance of GMAN degrades slowly when the predictions are made further into the future, and it performs well in the long-term (e.g., 12 time steps ahead) prediction, while still worse than STJGCN. In general, our model achieves the best performances at almost all horizons in terms of all three metrics on both datasets. 

\subsubsection{Model Size and Computation Time} \label{section4.3.5}

We present the comparison of model size and computation time of our STJGCN with graph-based baseline methods in Table~\ref{table7}. 

The results in four PeMS datasets demonstrate the high computation efficiency of our model. In terms of the model size, STJGCN has fewer parameters than most of the baseline models. In the training phase, our model runs faster than other methods except for STGCN. In the inference stage, STGCN runs very slowly as it adopts an iterative way to generate multi-step predictions, while STJGCN and Graph WaveNet are the most efficient. By further considering the prediction accuracy (see Table~\ref{table4}), our model shows superior ability in balancing predictive performances and time consumption as well as parameter settings.

The results in Seattle-Loop dataset show that out STJGCN compares favorably to baseline methods. In terms of the model size, STJGCN has not been affected by the larger amount of data, and still has fewer parameters than most of the baseline models. In the training phase, our STJGCN is faster than GMAN and Z-GCNETs. Other 6 baselines are more efficient than STJGCN but they show poor prediction performance (see Table~\ref{table4}). In the inference stage, STJGCN is only slower than STSGCN and AGCRN, while both of which have worse prediction accuracy than our model (see Table~\ref{table4}). 

\section{Conclusion} \label{Conclusion}

We proposed STJGCN, which models comprehensive and dynamic spatio-temporal correlations and aggregates multiple ranges of information to forecast the traffic conditions over several time steps ahead on a road network. When evaluated on five public traffic datasets, STJGCN showed high computation efficiency and outperformed 11 state-of-the-art baseline methods. Our model could be potentially
applied to other spatio-temporal data forecasting tasks, such as air quality inference and taxi demand prediction. We plan to investigate this as future works. 

\section*{Acknowledgment}

The research was supported by Natural Science Foundation of China (62272403, 61872306), and Fundamental Research Funds for the Central Universities (20720200031).

\ifCLASSOPTIONcaptionsoff
  \newpage
\fi

\bibliographystyle{IEEEtran}
\bibliography{STJGCN}

\begin{thebibliography}{10}
\providecommand{\url}[1]{#1}
\csname url@samestyle\endcsname
\providecommand{\newblock}{\relax}
\providecommand{\bibinfo}[2]{#2}
\providecommand{\BIBentrySTDinterwordspacing}{\spaceskip=0pt\relax}
\providecommand{\BIBentryALTinterwordstretchfactor}{4}
\providecommand{\BIBentryALTinterwordspacing}{\spaceskip=\fontdimen2\font plus
\BIBentryALTinterwordstretchfactor\fontdimen3\font minus
  \fontdimen4\font\relax}
\providecommand{\BIBforeignlanguage}[2]{{%
\expandafter\ifx\csname l@#1\endcsname\relax
\typeout{** WARNING: IEEEtran.bst: No hyphenation pattern has been}%
\typeout{** loaded for the language `#1'. Using the pattern for}%
\typeout{** the default language instead.}%
\else
\language=\csname l@#1\endcsname
\fi
#2}}
\providecommand{\BIBdecl}{\relax}
\BIBdecl

\bibitem{Wang-et-al:TKDE2020}
S.~Wang, J.~Cao, and P.~S. Yu, ``Deep learning for spatio-temporal data mining:
  A survey,'' \emph{IEEE Transactions on Knowledge and Data Engineering}, 2020.

\bibitem{Tedjopurnomo-et-al:TKDE2020}
D.~A. Tedjopurnomo, Z.~Bao, B.~Zheng, F.~M. Choudhury, and K.~Qin, ``A survey
  on modern deep neural network for traffic prediction: Trends, methods and
  challenges,'' \emph{IEEE Transactions on Knowledge and Data Engineering},
  2020.

\bibitem{Zheng-et-al:TITS2021}
C.~Zheng, C.~Wang, X.~Fan, J.~Qi, and X.~Yan, ``Stpc-net: Learn massive
  geo-sensory data as spatio-temporal point clouds,'' \emph{IEEE Transactions
  on Intelligent Transportation Systems}, 2021.

\bibitem{Li-et-al:ICLR2018}
Y.~Li, R.~Yu, C.~Shahabi, and Y.~Liu, ``Diffusion convolutional recurrent
  neural network: Data-driven traffic forecasting,'' in \emph{ICLR}, 2018.

\bibitem{Cheng-et-al:AAAI2018}
W.~Cheng, Y.~Shen, Y.~Zhu, and L.~Huang, ``A neural attention model for urban
  air quality inference: learning the weights of monitoring stations,'' in
  \emph{AAAI}, 2018, pp. 2151--2158.

\bibitem{Yu-et-al:IJCAI2018}
B.~Yu, H.~Yin, and Z.~Zhu, ``Spatio-temporal graph convolutional networks: A
  deep learning framework for traffic forecasting,'' in \emph{IJCAI}, 2018, pp.
  3634--3640.

\bibitem{Wu-et-al:IJCAI2019}
Z.~Wu, S.~Pan, G.~Long, J.~Jiang, and C.~Zhang, ``Graph wavenet for deep
  spatial-temporal graph modeling,'' in \emph{IJCAI}, 2019.

\bibitem{Guo-et-al:AAAI2019}
S.~Guo, Y.~Lin, N.~Feng, C.~Song, and HuaiyuWan, ``Attention based
  spatial-temporal graph convolutional networks for traffic flow forecasting,''
  in \emph{AAAI}, 2019, pp. 922--929.

\bibitem{Song-et-al:AAAI2020}
C.~Song, Y.~Lin, S.~Guo, and HuaiyuWan, ``Spatial-temporal synchronous graph
  convolutional networks: A new framework for spatial-temporal network data
  forecasting,'' in \emph{AAAI}, 2020.

\bibitem{Bai-et-al:NIPS2020}
L.~Bai, L.~Yao, C.~Li, X.~Wang, and C.~Wang, ``Adaptive graph convolutional
  recurrent network for traffic forecasting,'' in \emph{NeurIPS}, 2020.

\bibitem{Chen-et-al:ICML2021}
Y.~Chen, I.~Segovia-Dominguez, and Y.~R. Gel, ``Z-gcnets: Time zigzags at graph
  convolutional networks for time series forecasting,'' in \emph{ICML}, 2021.

\bibitem{Wu-et-al:TNNLS2021}
Z.~Wu, S.~Pan, F.~Chen, G.~Long, C.~Zhang, and P.~S. Yu, ``A comprehensive
  survey on graph neural networks,'' \emph{IEEE Transactions on Neural Networks
  and Learning Systems}, vol.~32, no.~1, pp. 4--24, 2021.

\bibitem{Kipf-and-Welling:ICLR2017}
T.~N. Kipf and M.~Welling, ``Semi-supervised classification with graph
  convolutional networks,'' in \emph{ICLR}, 2017.

\bibitem{Zhang-et-al:NIPS2018}
M.~Zhang and Y.~Chen, ``Link prediction based on graph neural networks,'' in
  \emph{NeurIPS}, 2018, pp. 5165--5175.

\bibitem{Bronstein-et-al:SPM2017}
M.~M. Bronstein, J.~Bruna, Y.~LeCun, A.~Szlam, and P.~Vandergheynst,
  ``Geometric deep learning: going beyond euclidean data,'' \emph{IEEE Signal
  Processing Magazine}, vol.~34, no.~4, pp. 18--42, 2017.

\bibitem{Niepert-et-al:ICML2016}
M.~Niepert, M.~Ahmed, and K.~Kutzkov, ``Learning convolutional neural networks
  for graphs,'' in \emph{ICML}, 2016, pp. 2014--2023.

\bibitem{Hamilton-et-al:NIPS2017}
W.~L. Hamilton, R.~Ying, and J.~Leskovec, ``Inductive representation learning
  on large graphs,'' in \emph{NeurIPS}, 2017, pp. 1024--1034.

\bibitem{Velickovic-et-al:ICLR2018}
P.~Veličković, G.~Cucurull, A.~Casanova, A.~Romero, P.~Liò, and Y.~Bengio,
  ``Graph attention networks,'' in \emph{ICLR}, 2018.

\bibitem{jia2020residual}
J.~Jia and A.~R. Benson, ``Residual correlation in graph neural network
  regression,'' in \emph{Proceedings of the 26th ACM SIGKDD International
  Conference on Knowledge Discovery \& Data Mining}, 2020, pp. 588--598.

\bibitem{Li-et-al:AAAI2020}
Q.~Li, Z.~Han, and X.-M. Wu, ``Deeper insights into graph convolutional
  networks for semi-supervised learning,'' in \emph{AAAI}, 2020.

\bibitem{Bruna-et-al:ICLR2014}
J.~Bruna, W.~Zaremba, A.~Szlam, and Y.~LeCun, ``Spectral networks and deep
  locally connected networks on graphs,'' in \emph{ICLR}, 2014.

\bibitem{Defferrard-et-al:NIPS2016}
M.~Defferrard, X.~Bresson, and P.~Vandergheynst, ``Convolutional neural
  networks on graphs with fast localized spectral filtering,'' in
  \emph{NeurIPS}, 2016, pp. 3844--3852.

\bibitem{Zhang-et-al:TITS2021}
S.~Zhang, Y.~Guo, P.~Zhao, C.~Zheng, and X.~Chen, ``A graph-based temporal
  attention framework for multi-sensor traffic flow forecasting,'' \emph{IEEE
  Transactions on Intelligent Transportation Systems}, 2021.

\bibitem{Yan-et-al:AAAI2018}
S.~Yan, Y.~Xiong, and D.~Lin, ``Spatial temporal graph convolutional networks
  for skeleton-based action recognition,'' in \emph{AAAI}, 2018, p.
  3482–3489.

\bibitem{Shi-et-al:CVPR2019}
L.~Shi, Y.~Zhang, J.~Cheng, and H.~Lu, ``Two-stream adaptive graph
  convolutional networks for skeleton-based action recognition,'' in
  \emph{CVPR}, 2019, p. 12026–12035.

\bibitem{Yin-et-al:arXiv2020}
X.~Yin, G.~Wu, J.~Wei, Y.~Shen, H.~Qi, and B.~Yin, ``A comprehensive survey on
  traffic prediction,'' \emph{arXiv preprint arXiv:2004.08555}, 2020.

\bibitem{Sun-et-al:TKDE2020}
J.~Sun, J.~Zhang, Q.~Li, X.~Yi, Y.~Liang, and Y.~Zheng, ``Predicting citywide
  crowd flows in irregular regions using multi-view graph convolutional
  networks,'' \emph{IEEE Transactions on Knowledge and Data Engineering}, 2020.

\bibitem{Gu-et-al:TKDE2020}
J.~Gu, Q.~Zhou, J.~Yang, Y.~Liu, F.~Zhuang, Y.~Zhao, and H.~Xiong, ``Exploiting
  interpretable patterns for flow prediction in dockless bike sharing
  systems,'' \emph{IEEE Transactions on Knowledge and Data Engineering}, 2020.

\bibitem{Jiang-et-al:TKDE2021}
R.~Jiang, Z.~Cai, Z.~Wang, C.~Yang, Z.~Fan, Q.~Chen, K.~Tsubouchi, X.~Song, and
  R.~Shibasaki, ``Deepcrowd: A deep model for large-scale citywide crowd
  density and flow prediction,'' \emph{IEEE Transactions on Knowledge and Data
  Engineering}, 2021.

\bibitem{Guo-et-al:TKDE2021}
S.~Guo, Y.~Lin, H.~Wan, X.~Li, and G.~Cong, ``Learning dynamics and
  heterogeneity of spatial-temporal graph data for traffic forecasting,''
  \emph{IEEE Transactions on Knowledge and Data Engineering}, 2021.

\bibitem{Ma-et-al:TRC2015}
X.~Ma, Z.~Tao, Y.~Wang, H.~Yu, and Y.~Wang, ``Long short-term memory neural
  network for traffic speed prediction using remote microwave sensor data,''
  \emph{Transportation Research Part C: Emerging Technologies}, vol.~54, pp.
  187--197, 2015.

\bibitem{Zhang-et-al:AAAI2017}
J.~Zhang, Y.~Zheng, and D.~Qi, ``Deep spatio-temporal residual networks for
  citywide crowd flows prediction,'' in \emph{AAAI}, 2017, pp. 1655--1661.

\bibitem{Yao-et-al:AAAI2018}
H.~Yao, F.~Wu, J.~Ke, X.~Tang, Y.~Jia, S.~Lu, P.~Gong, J.~Ye, and Z.~Li, ``Deep
  multi-view spatial-temporal network for taxi demand prediction,'' in
  \emph{AAAI}, 2018, pp. 2588--2595.

\bibitem{Yao-et-al:AAAI2019}
H.~Yao, X.~Tang, H.~Wei, G.~Zheng, and Z.~Li, ``Revisiting spatial-temporal
  similarity: A deep learning framework for traffic prediction,'' in
  \emph{AAAI}, 2019.

\bibitem{Zheng-et-al:TITS2020}
C.~Zheng, X.~Fan, C.~Wen, L.~Chen, C.~Wang, and J.~Li, ``Deepstd: Mining
  spatio-temporal disturbances of multiple context factors for citywide traffic
  flow prediction,'' \emph{IEEE Transactions on Intelligent Transportation
  Systems}, vol.~21, no.~9, pp. 3744--3755, 2020.

\bibitem{Zhang-et-al:TKDE2020}
J.~Zhang, Y.~Zheng, J.~Sun, and D.~Qi, ``Flow prediction in spatio-temporal
  networks based on multitask deep learning,'' \emph{IEEE Transactions on
  Knowledge and Data Engineering}, vol.~32, no.~3, 2020.

\bibitem{Ye-et-al:arXiv2020}
J.~Ye, J.~Zhao, K.~Ye, and C.~Xu, ``How to build a graph-based deep learning
  architecture in traffic domain: A survey,'' \emph{arXiv preprint
  arXiv:2005.11691}, 2020.

\bibitem{lau2021spatio}
Y.~H. Lau and R.~C.-W. Wong, ``Spatio-temporal graph convolutional networks for
  traffic forecasting: Spatial layers first or temporal layers first?'' in
  \emph{Proceedings of the 29th International Conference on Advances in
  Geographic Information Systems}, 2021, pp. 427--430.

\bibitem{Zheng-et-al:AAAI2020}
C.~Zheng, X.~Fan, C.~Wang, and J.~Qi, ``Gman: A graph multi-attention network
  for traffic prediction,'' in \emph{AAAI}, 2020, pp. 1234--1241.

\bibitem{Wang-et-al:WWW2020}
X.~Wang, Y.~Ma, Y.~Wang, W.~Jin, X.~Wang, J.~Tang, C.~Jia, and J.~Yu, ``Traffic
  flow prediction via spatial temporal graph neural network,'' in \emph{WWW},
  2020, pp. 1082--1092.

\bibitem{Vaswani-et-al:NIPS2017}
A.~Vaswani, N.~Shazeer, N.~Parmar, J.~Uszkoreit, L.~Jones, A.~N. Gomez, Łukasz
  Kaiser, and I.~Polosukhin, ``Attention is all you need,'' in \emph{NeurIPS},
  2017, pp. 5998--6008.

\bibitem{lu2020agstn}
Y.-J. Lu and C.-T. Li, ``Agstn: Learning attention-adjusted graph
  spatio-temporal networks for short-term urban sensor value forecasting,'' in
  \emph{2020 IEEE International Conference on Data Mining (ICDM)}.\hskip 1em
  plus 0.5em minus 0.4em\relax IEEE, 2020, pp. 1148--1153.

\bibitem{Wang-et-al:KDD2019}
Y.~Wang, H.~Yin, H.~Chen, T.~Wo, J.~Xu, and K.~Zheng, ``Origin-destination
  matrix prediction via graph convolution: a new perspective of passenger
  demand modeling,'' in \emph{KDD}, 2019, pp. 1227--1235.

\bibitem{Chen-et-al:AAAI2020}
W.~Chen, L.~Chen, Y.~Xie, W.~Cao, Y.~Gao, and X.~Feng, ``Multi-range attentive
  bicomponent graph convolutional network for traffic forecasting,'' in
  \emph{AAAI}, 2020.

\bibitem{Deng-et-al:KDD2016}
D.~Deng, C.~Shahabi, U.~Demiryurek, L.~Zhu, R.~Yu, and Y.~Liu, ``Latent space
  model for road networks to predict time-varying traffic,'' in \emph{KDD},
  2016.

\bibitem{Nair-and-Hinton:ICML2010}
V.~Nair and G.~E. Hinton, ``Rectified linear units improve restricted boltzmann
  machines,'' in \emph{ICML}, 2010, pp. 807--814.

\bibitem{Schuster-and-Paliwal:TSP1997}
M.~Schuster and K.~K. Paliwal, ``Bidirectional recurrent neural networks,''
  \emph{IEEE Transactions on Signal Processing}, vol.~45, no.~11, pp.
  2673--2681, 1997.

\bibitem{Oord-et-al:arXiv2016}
A.~van~den Oord, S.~Dieleman, H.~Zen, K.~Simonyan, O.~Vinyals, A.~Graves,
  N.~Kalchbrenner, A.~Senior, and K.~Kavukcuoglu, ``Wavenet: A generative model
  for raw audio,'' \emph{arXiv preprint arXiv:1609.03499}, 2016.

\bibitem{Bai-et-al:arXiv2018}
S.~Bai, J.~Z. Kolter, and V.~Koltun, ``An empirical evaluation of generic
  convolutional and recurrent networks for sequence modeling,'' \emph{arXiv
  preprint arXiv:1803.01271}, 2018.

\bibitem{He-et-al:CVPR2016}
K.~He, X.~Zhang, S.~Ren, and J.~Sun, ``Deep residual learning for image
  recognition,'' in \emph{CVPR}, 2016, pp. 770--778.

\bibitem{cui2018deep}
Z.~Cui, R.~Ke, and Y.~Wang, ``Deep bidirectional and unidirectional lstm
  recurrent neural network for network-wide traffic speed prediction,''
  \emph{arXiv preprint arXiv:1801.02143}, 2018.

\bibitem{cui2019traffic}
Z.~Cui, K.~Henrickson, R.~Ke, and Y.~Wang, ``Traffic graph convolutional
  recurrent neural network: A deep learning framework for network-scale traffic
  learning and forecasting,'' \emph{IEEE Transactions on Intelligent
  Transportation Systems}, 2019.

\bibitem{shin2022pgcn}
Y.~Shin and Y.~Yoon, ``Pgcn: Progressive graph convolutional networks for
  spatial-temporal traffic forecasting,'' \emph{arXiv preprint
  arXiv:2202.08982}, 2022.

\bibitem{Ioffe-and-Szegedy:ICML2015}
S.~Ioffe and C.~Szegedy, ``Batch normalization: Accelerating deep network
  training by reducing internal covariate shift,'' in \emph{ICML}, 2015, pp.
  448--456.

\bibitem{Kingma-and-Ba:ICLR2015}
D.~P. Kingma and J.~L. Ba, ``Adam: a method for stochastic optimization,'' in
  \emph{ICLR}, 2015.

\bibitem{VAR:1994}
J.~D. Hamilton, \emph{Time series analysis}.\hskip 1em plus 0.5em minus
  0.4em\relax Princeton university press, 1994.

\bibitem{Sutskever-et-al:NIPS2014}
I.~Sutskever, O.~Vinyals, and Q.~V. Le, ``Sequence to sequence learning with
  neural networks,'' in \emph{NeurIPS}, 2014, pp. 3104--3112.

\bibitem{Drucker-et-al:NIPS1997}
H.~Drucker, C.~J. Burges, L.~Kaufman, A.~Smola, and V.~Vapoik, ``Support vector
  regression machines,'' in \emph{NeurIPS}, 1997, pp. 155--161.

\bibitem{Carlsson-and-Silva:2010}
G.~Carlsson and V.~de~Silva, ``Zigzag persistence,'' \emph{Foundations of
  computational mathematics}, vol.~10, pp. 367--405, 2010.

\end{thebibliography}

\begin{IEEEbiography}[{\includegraphics[width = 1in,height = 1.25in, clip, keepaspectratio]{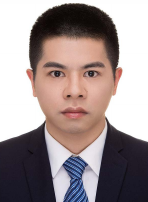}}]{Chuanpan Zheng}
	 received the B.Sc. degree in applied physics from Shandong University, Jinan, China, in 2012. He is currently working toward the Ph.D. degree in computer science and technology at Fujian Key Laboratory of Sensing and Computing for Smart Cities, School of Informatics, Xiamen University, China. His research interests include spatio-temporal data representation learning and graph neural networks.
\end{IEEEbiography}
\begin{IEEEbiography}[{\includegraphics[width = 1in, height = 1.25in, clip, keepaspectratio]{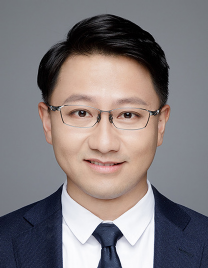}}]{Xiaoliang Fan}
	(M’06-SM’18) is a Senior Research Specialist at Fujian Key Laboratory of Sensing and Computing for Smart Cites, School of Informatics, and Key Laboratory of Multimedia Trusted Perception and Efficient Computing, Ministry of Education of China, Xiamen University, China. He received his PhD degree at University Pierre and Marie CURIE, France in 2012. His research interests include trustworthy AI and federated learning, spatio-temporal data mining, and services computing, etc. He has published 70+ journals (IEEE TSC/TMC/TITS, etc.) and top conferences (AAAI, IJCAI, WWW, etc.) papers. His works are funded by NSFC and many industry collaborators. Dr. FAN is an IEEE Senior Member, and CCF Senior Member.
\end{IEEEbiography}
\begin{IEEEbiography}[{\includegraphics[width = 1in, height = 1.25in, clip, keepaspectratio]{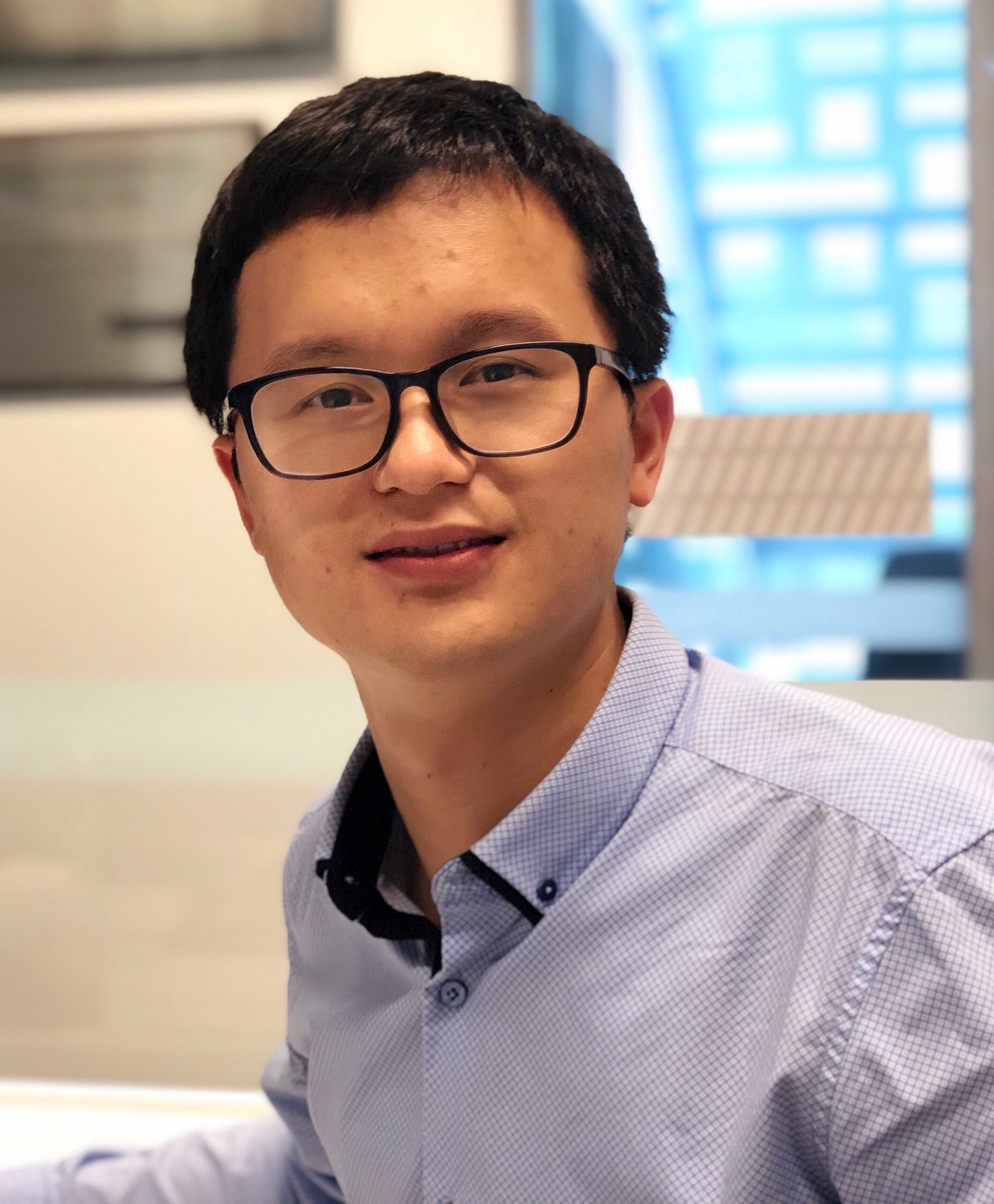}}]{Shirui Pan} 
	received a Ph.D. in computer science from the University of Technology Sydney (UTS), Ultimo, NSW, Australia. He is currently a full professor with School of Information and Communication Technology, Griffith University, Australia.  His research interests include data mining and machine learning. To date, Dr. Pan has published over 100 research papers in top-tier journals and conferences, including the TPAMI, TKDE,  TNNLS, NeurIPS, ICML, and KDD. He is recognised as one of the \textit{AI 2000 AAAI/IJCAI Most Influential Scholars} in Australia (2021).
\end{IEEEbiography}
\begin{IEEEbiography}[{\includegraphics[width = 1in, height = 1.25in, clip, keepaspectratio]{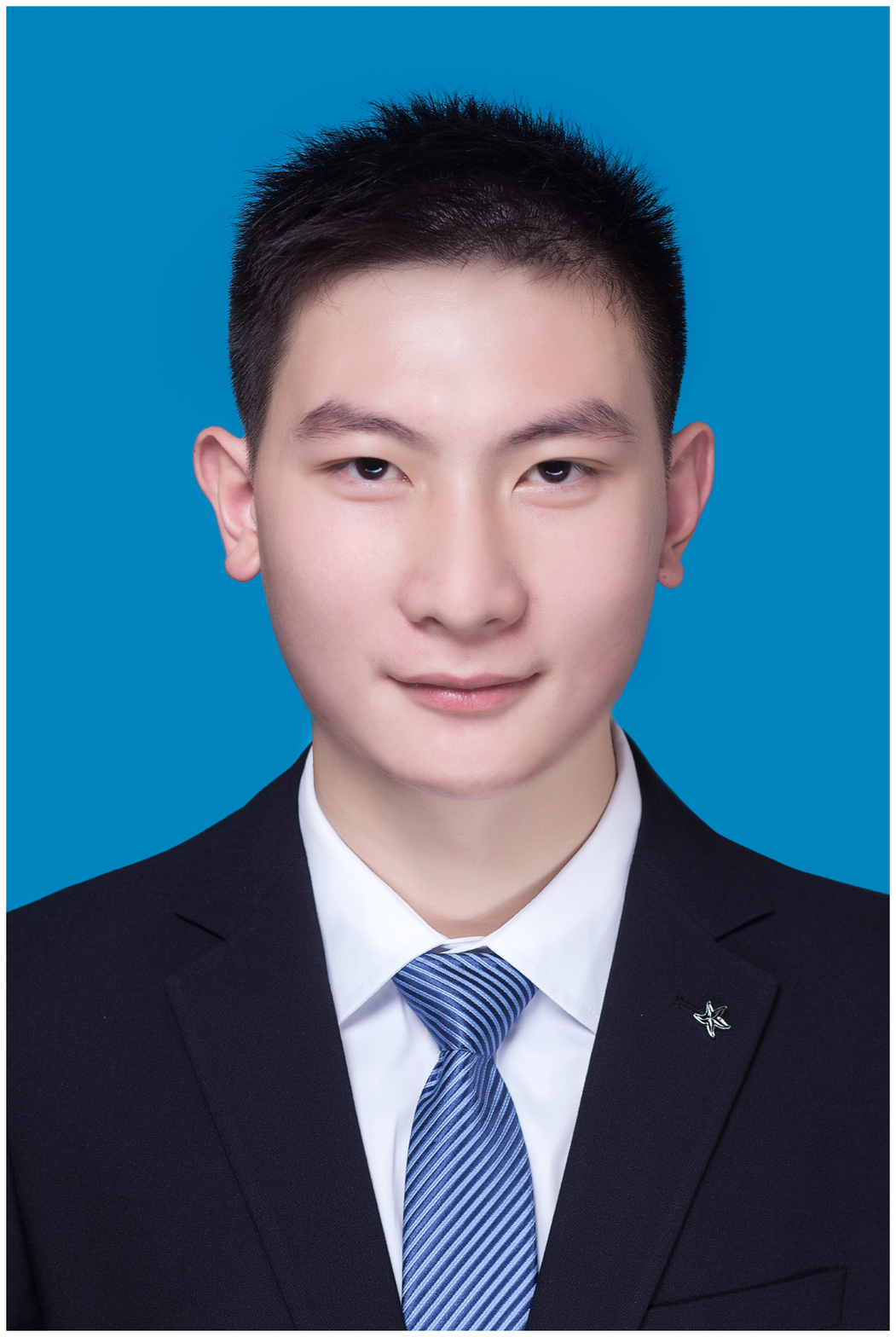}}]{Haibing Jin} 
	  received the B.Sc. degree in Computer Science and Technology from Huazhong Agricultural University, Wuhan, China, in 2022. He is currently working toward the MA.Eng. degree in computer science and technology at Fujian Key Laboratory of Sensing and Computing for Smart Cities, School of Informatics, Xiamen University, China. His research interests include spatio-temporal data representation learning and federated learning.
\end{IEEEbiography}
\begin{IEEEbiography}[{\includegraphics[width = 1in, height = 1.25in, clip, keepaspectratio]{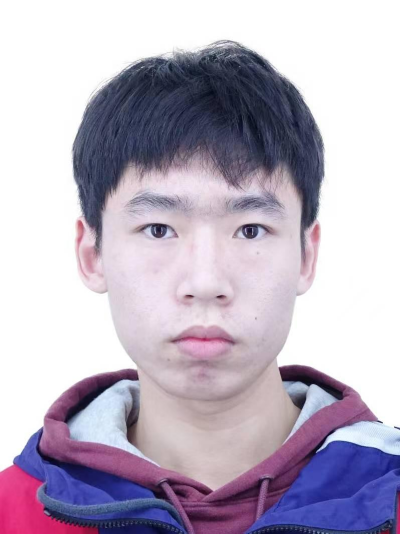}}]{Zhaopeng Peng} 
	  received the B.Sc. degree in Digital media technology from Shandong University, Weihai, China, in 2022. He is currently working toward the MA.Eng. degree in computer science and technology at Fujian Key Laboratory of Sensing and Computing for Smart Cities, School of Informatics, Xiamen University, China. His research interests include spatio-temporal data representation learning and federated learning.
\end{IEEEbiography}
\begin{IEEEbiography}[{\includegraphics[width = 1in, height = 1.25in, clip, keepaspectratio]{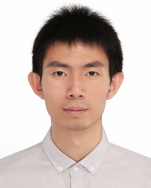}}]{Zonghan Wu} 
	obtained a B.S. in systems science from the University of Shanghai for Science and Technology. He received a M.S. in statistics from the Link{\"o}ping University. He is now pursuing a Ph.D. in computer science from the University of Technology Sydney (UTS), Ultimo, NSW, Australia.	
\end{IEEEbiography}
\begin{IEEEbiography}[{\includegraphics[width = 1in, height = 1.25in, clip, keepaspectratio]{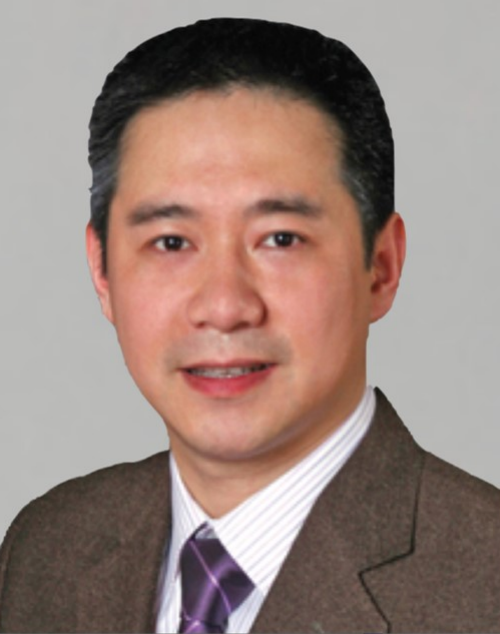}}]{Cheng Wang} (M'04-SM'16)
	received the Ph.D. degree in information and communication engineering from National University of Defense Technology, Changsha, China, in 2002. He is currently a NanQiang Professor with School of Informatics, and Director of Fujian Key Laboratory of Sensing and Computing for Smart Cities, both at Xiamen University, China. His research interests include remote sensing image processing, mobile LiDAR data analysis, and multi-sensor fusion. He has co-authored over 150 papers in referred journals and top conferences including IEEE-TGRS, PR, IEEE-TITS, AAAI, CVPR, IJCAI, and ISPRS-JPRS.
\end{IEEEbiography}
\begin{IEEEbiography}[{\includegraphics[width = 1in, height = 1.25in, clip, keepaspectratio]{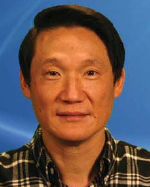}}]{Philip S. Yu} (F'93)
	received the Ph.D. degree in electrical engineering from Stanford University, Stanford, CA, USA. He is currently a Distinguished Professor of computer science with the University of Illinois at Chicago, Chicago, IL, USA, where he is also the Wexler Chair in Information Technology. He has published more than 830 articles in refereed journals and conferences. He holds or has applied for more than 300 U.S. patents. His research interests include big data, data mining, data streams, databases, and privacy.
	
	Dr. Yu is a fellow of the ACM. He received the ACM SIGKDD 2016 Innovation Award, the Research Contributions Award from the IEEE International Conference on Data Mining in 2003, and the Technical Achievement Award from the IEEE Computer Society in 2013.
\end{IEEEbiography}





\end{document}